\documentclass[journal]{IEEEtran}

\usepackage{booktabs}
\usepackage{comment}
\usepackage{float}
\usepackage{multirow}%
\usepackage{amsmath,amssymb,amsfonts}%
\usepackage{algorithmicx}%
\usepackage{algpseudocode}%
\usepackage{listings}%
\usepackage{caption}
\usepackage{subfigure}
\usepackage{graphicx}

\usepackage[outdir=./]{epstopdf}

\usepackage{comment}
\usepackage{algorithm}

\usepackage[figuresright]{rotating}

\usepackage[numbers]{natbib}
\algrenewcommand\algorithmicrequire{\textbf{Input:}}
\algrenewcommand\algorithmicensure{\textbf{Output:}}

\begin{document}
\title{Multiplayer Battle Game-Inspired Optimizer for Complex Optimization Problems}
\author{Yuefeng XU,  Rui ZHONG, Chao ZHANG, Jun YU
\thanks{C. ZHANG and Y. Xu are with the University of Fukui, Japan.}
\thanks{R. ZHONG is with the Hokkaido University, Japan.}
\thanks{J. Yu is with the Niigata University, Japan.}}

\maketitle

\begin{abstract}
 Various popular multiplayer battle royale games share a lot of common elements. Drawing from our observations, we summarized these shared characteristics and subsequently proposed a novel heuristic algorithm named multiplayer battle game-inspired optimizer (MBGO). The proposed MBGO streamlines mainstream multiplayer battle royale games into two discrete phases: movement and battle. Specifically, the movement phase incorporates the principles of commonly encountered ``safe zones'' to incentivize participants to relocate to areas with a higher survival potential. The battle phase simulates a range of strategies adopted by players in various situations to enhance the diversity of the population. To evaluate and analyze the performance of the proposed MBGO, we executed it alongside eight other algorithms, including three classics and five latest ones, across multiple diverse dimensions within the CEC2017 and CEC2020 benchmark functions. In addition, we employed several industrial design problems to evaluate the scalability and practicality of the proposed MBGO. The results of the statistical analysis reveal that the novel MBGO demonstrates significant competitiveness, excelling not only in convergence speed, but also in achieving high levels of convergence accuracy across both benchmark functions and real-world problems.

\end{abstract}


\section{Introduction}
\label{sec1}
Optimization problems \cite{Optimization} are a frequently encountered but highly important class of topics in the field of computer science and math, aiming to find a feasible global optimum within acceptable costs under given constraints. Thanks to the continuous dedication and improvement of practitioners, some classic optimization algorithms, including Newton's method \cite{NewtonMethod}, linear programming \cite{LinearPro}, and hill-climbing methods \cite{hill-climbing}, have been successfully applied to various industrial scenarios. However, as the dimensionality of the problem increases, and the complexity of mathematical modeling intensifies, these aforementioned algorithms also encounter challenges in addressing increasingly complex problems. These dilemmas force practitioners to find ways to improve the performance of optimization algorithms\cite{optimization_algorithms}.

Evolutionary computation (EC), as a novel optimization approach, has rapidly garnered interest thanks to its outstanding characteristics, such as problem independence, high robustness, and parallel processing capabilities \cite{EC2006, EC1997}. As a pioneer member of the EC algorithms, genetic algorithm (GA) incrementally enhances the precision of candidate solutions by emulating the principle of ``survival of the fittest'' observed in biological evolution, ultimately converging towards the global optimum \cite{GA1994, GA2001}. This concept, borrowing population-based iterative optimization ideas, has provided fresh inspiration to a multitude of practitioners, catalyzing substantial growth in the field of EC since the 1990s \cite{EChis1997}. Up to the present, EC algorithms have been successfully solving various optimization problems featuring distinct characteristics, including multi-modal optimization \cite{model2019, yu2019}, multi-objective optimization \cite{objective2015, objective2021}, dynamic optimization \cite{dynamic2011, dynamic2012}, and others \cite{other2011, yu201905, other2022}.

The widespread adoption of EC algorithms has led to the emergence and dissemination of numerous algorithms beyond the classic GA. For example, particle swarm optimization (PSO) treats each candidate solution as an individual bird and updates candidate solutions by simulating the foraging behavior of a flock of birds \cite{pso}. Differential evolution (DE) leverages differential information between individuals to efficiently generate new candidate solutions and converge rapidly toward the global optimum \cite{DE}. Compared with the above two algorithms, the whale optimization algorithm (WOA) primarily emulates the hunting behavior of humpback whales and does not necessitate additional parameter settings  \cite{WOA}. In addition to these well-established algorithms, many interesting new algorithms are developed every year, such as honey badger algorithm (HBA) \cite{HBA}, tunicate swarm algorithm (TSA) \cite{TSA}, the sailfish optimization algorithm (SFO) \cite{SFO}, seagull optimization algorithm (SOA) \cite{SOA}, and aquila optimizer (AO) \cite{AO}. After decades of development, the EC community has seen the inclusion of hundreds of novel algorithms. While the motivations behind these algorithms may vary, they can still be broadly categorized into six major categories, namely bio-inspired, plant-based, swarm-based, human-inspired, physics-based, and math-based.

To the best of our knowledge, only a limited number of researchers draw inspiration from games and propose a handful of algorithms, even though games have been an integral part of human life since ancient times. Especially with the emergence of e-sports games, gaming provides individuals with the opportunity to relish the excitement of competition and kindles an interest in communication. Thus, we are not constrained to a particular game but aim to distill the common patterns shared by these games that captivate individuals, and subsequently, design a new optimization algorithm referred to as multiplayer battle game-inspired optimizer (MBGO). The proposed MBGO refines the competitive process of battle games into two distinct phases, labeled the movement phase and the battle phase. Through the repeated execution of these two stages, a balance between exploration and exploitation is achieved, resulting in the proposed MBGO displaying remarkable performance. Here, we summarize the contributions of the new MBGO algorithm as follows:

\begin{itemize}
      \item A new game-inspired optimization algorithm is crafted to achieve a balance between exploration and exploitation from a novel perspective. This design enables the new MBGO algorithm to achieve swift convergence while maintaining a high level of population diversity.
      \item The movement phase uses the distribution information of the population to partition the entire search space into two different areas. Individuals located in these separate areas employ different strategies to update their positions to converge toward potential areas quickly.
      \item The battle phase uses individual fitness information instead of Euclidean distance to simulate the battle strategies adopted by individuals under various psychological conditions when encountering opponents, which is focused on maintaining the diversity of the population.
      \item A series of analytical experiments are conducted to investigate the performance of the new MBGO. The experimental results demonstrate that the proposed MBGO exhibits robust competitiveness and significant potential, consistently ranking among the top performers in both test function sets and real-world industrial problems.
\end{itemize}

The structure of the remaining sections is outlined as follows. Sec.\ref{sec2} summarizes related studies and highlights the distinctions between our work and them. Sec.\ref{sec3} provides the motivation and presents a detailed description of the mathematical model of the proposed MBGO. A large number of comparative experiments are conducted to evaluate the performance of the proposed MBGO in Sec.\ref{sec4}. We further analyze the principles of effectiveness and scalability of the proposed MBGO and conclude our work in Sec.\ref{sec5} and Sec.\ref{sec6}.

\section{Related work}
\label{sec2}

We conducted a comprehensive search across multiple prominent databases and found the fact that there is a clear lack of meta-heuristics algorithms rooted in the principles and mechanics of games. This discrepancy highlights a unique domain teeming with innovative potential and unexplored opportunities. Specifically, the utilization of game-inspired concepts holds the potential to yield pioneering and highly effective optimization techniques. Given the trend of interdisciplinary cross-pollination\cite{grodal2008cross}, it is foreseeable that an increasing number of researchers and developers will be drawn to the intriguing intersection of games and optimization \cite{Game}. Thus, it is time to broaden our perspectives, directing our efforts not only toward the creation of innovative and captivating games but also toward the exploration of novel avenues for the application of game-derived inspiration in diverse fields. This is also one of the driving forces that prompted the emergence of this work. In the subsequent sections, we will introduce two papers that fall within the same category as this one and delineate the distinctions between them.

As the first game-inspired EC algorithm, the battle royale optimization algorithm (BRO) \cite{BRO} mainly draws inspiration from the ``battle royale'' genre within digital gaming. It emulates the player's survival strategies to gradually converge towards the optimal area. Since this type of game includes renowned titles such as Player Unknown's Battlegrounds (PUBG)\cite{PUBGresearch} and Apex Legends, the foundational BRO selects the ``deathmatch'' of PUBG, simulating the battle behaviors of multiple players to address single-objective optimization problems. Shortly after the introduction of the basic BRO, several enhanced versions were also proposed. For example, Akan et al. introduced an additional movement operator to further improve the search efficiency \cite{Akan2022}, and they extended the MBGO algorithm to tackle discrete optimization problems \cite{MBGObin}. The advent of these variants further underscores the potential for the development of EC algorithms rooted in the world of gaming.

The other related study within the game-inspired category, the squid game optimizer (SGO) \cite{SGO}, was recently introduced. The SGO takes inspiration from the fundamental rules of a traditional Korean game to design optimization operators. In this game, multiple players are divided into two distinct teams. One team serves as the offensive players, while the other takes on the role of defensive players, all within a playing field shaped like a squid. The primary objectives within this game involve the attackers completing their objectives or the teams eliminating each other. Drawing from these game principles, SGO employs a unique approach in which individuals are categorized into two distinct groups, each with its own distinct evolutionary strategies. This innovative concept aligns with the dynamics of the game and offers a fresh perspective for optimization algorithms. Since the SGO was proposed not long ago, there are not many improvements and applications yet. 

These similar studies persist in their focus on a specific game, devising new optimization frameworks through the simplification of gameplay\cite{fabricatore2007gameplay}. Achieving a lossless representation of a game is highly challenging. Thus, it is crucial to select the operations that need to be retained during the simplification process, which often requires a robust background in optimization design. Because of the above considerations, we sought to explore a distinctive pathway for algorithmic design. Consequently, instead of fixating on a specific game, we distill the common features inherent in battle games to construct a brand-new optimization framework. Furthermore, our approach extends beyond the game mechanics by incorporating player behaviour\cite{pierce2017behavior}, including the mentalities and strategies employed when facing different enemies, into the new algorithm for the first time. This integration aims to simulate a dynamic interaction between the game and multiple players, under the belief that a game becomes truly interesting and charming only when players actively engage in it. Finally, we present a new MBGO algorithm founded on the characteristics of simulated battle games and player behaviors. 
 
\section{Multiplayer Battle Game-Inspired Optimizer}
\label{sec3}

\subsection{Motivation}
\label{sec3.1}

Battle royale games\cite{BGgameplay,battleground}, such as PUBG, Infinite Law, Warzone, and Call of Duty, typically integrate elements from both survival games, focusing on exploration and equipment collection, and competitive gaming, where the objective is to eliminate opponents until only one person remains. While the rules of these games may not be identical, there are discernible common patterns that characterize their gameplay. For example, a prevalent game mechanic, known as ``safe zones'', is widely employed to compel players into more confined areas, fostering encounters and battles among them. Over time, the designated safe area progressively diminishes, and players situated outside this zone face the risk of injury or elimination until only the last individual or team remains as the victor. We acknowledge the crucial role played by the``safe zone'' mechanic in these games, significantly shaping player movement and battle strategies. Recognizing the widespread influence and impact of this mechanic, we have intentionally directed our simulation efforts toward the generation and utilization of the safe zone.

Another factor that affects the outcome of a match is the players themselves, as different players tend to choose different strategies when faced with even the same situation. For example, cautious players typically opt to engage in battle when they have a higher likelihood of success and avoid confrontation when the odds are less favorable. On the other hand, aggressive players consistently lean towards engaging in battle upon encountering a new opponent. Furthermore, the quality of equipment significantly influences players' decision-making. Generally, superior equipment correlates with an elevated probability of winning, instilling greater confidence in players when engaging in battle. To reflect the importance of players, when designing the new algorithm, we especially introduce several different strategies to simulate the behaviors of players under different psychological conditions. Our intention is for these strategies to effectively preserve individual diversity.

Since our primary objective is the development of an effective algorithm rather than blindly pursuing simulation games, we are open to making trade-offs on the rules. Here, we consider each player as a candidate solution within the search space and map their actions to coordinate updates in the solution space. As players are gradually eliminated, the number of individuals decreases in the game. However, to maintain a constant population size, we have chosen not to adopt this rule. Furthermore, we overlook equipment differences between players and instead treat each individual equally. This implies adopting different movement and battle strategies with equal probability for all individuals. In the subsequent subsection, we provide a comprehensive overview of the proposed MBGO algorithm.

\subsection{Mathematical model}
\label{sec3.2}

\begin{figure}[htbp]
\centering
\includegraphics[scale=0.5]{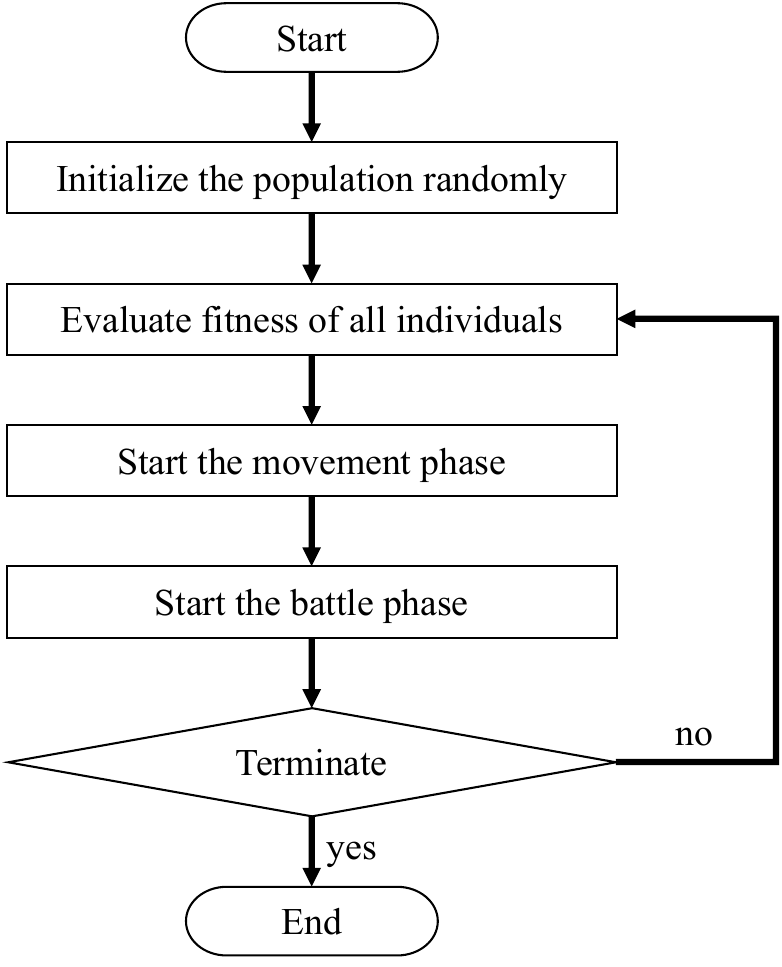}
\caption{The universal optimization framework of proposed MBGO algorithm.}
\label{fig:img1}
\end{figure}

The newly proposed MBGO algorithm is a population-based heuristic algorithm that iteratively refines the population by simulating common mechanisms found in battle games and behaviours taken by players under various psychological states. Similar to typical EC algorithms, the MBGO first randomly generates multiple individuals to form an initial population. Then, the Euclidean distance between the best individual and the worst individual in the current population is employed to define the ``safe zone'', effectively partitioning the entire search space into the safe area and the non-safe area. Individuals located in different areas use different movement strategies to generate new individuals. Importantly, individuals choose to update their current positions only if the new individual represents an improvement; otherwise, the current individual remains unchanged. After all individuals have gone through this movement phase, they transition into the battle phase. During this stage, each individual randomly chooses another individual as an opponent and determines the battle strategy to generate a new individual. The proposed MBGO still maintains the use of elite selection in the battle phase, accepting the new individual only if it is superior; otherwise, the individual ``loses'' the battle and is reborn in the same place. Subsequently, individuals re-enter the movement and battle phases again, and this cycle continues iteratively until the termination condition is met. Finally, Fig. \ref{fig:img1} provides a simple optimization process of the proposed MBGO algorithm, which consists of initialization, movement phase, and battle phase.

Before delving into the detailed introduction of three core operators, we first define some symbols that will be used in the following to avoid ambiguity. Here, we take the minimization problem as an example, assuming that the optimization problem contains $D$ variables, that is, a $D$-dimensional problem. Thus, for any individual $\mathbf{x}_i$ (i.e., candidate solution), it comprises $D$ variables, represented as $\mathbf{x}_i=(x_{i}^1, x_{i}^2, ..., x_{i}^D)$. For any dimension $k  (k \in \{1, 2, ..., D\})$, its upper bound and lower bound are defined as $UB^k$ and $LB^k$ respectively. Additionally, $\mathbf{x}_{best}$ and $\mathbf{x}_{worst}$ represent the best individual and the worst individual in the current population, and $Fit()$ is the evaluation function and returns the fitness value of the passed-in individual.

\textbf{Initialization} is executed only once to generate the initial population. Among various initialization methods, we employ the most common approach: random initialization. For any individual $\mathbf{x}_i$, Eq.(\ref{eqn:1}) uses the uniform distribution to generate the initial value in the $k$-th dimension:

\begin{equation}
\label{eqn:1}
x_{i}^{k}= LB_{i}^{k} + rand(0,1) \times (UB_{i}^{k} - LB_{i}^{k})
\end{equation}

\noindent where $rand(0,1)$ satisfies the uniform distribution and returns a random number between 0 and 1.

Suppose the population size, a hyperparameter that needs to be defined in advance, is set to $N$, the initial population can be represented in the form of a matrix, as follows:

\begin{equation}
\begin{split}\text{The initial population} = 
\begin{bmatrix}  
  x_{1}^1 & x_{1}^2 & x_{1}^3 & \cdots & x_{1}^D \\  
  x_{2}^1 & x_{2}^2 & x_{2}^3 & \cdots & x_{2}^D \\  
  \vdots & \vdots & \vdots & \vdots \\  
  x_{N}^1 & x_{N}^2 & x_{N}^3 & \cdots & x_{N}^D  \nonumber
\end{bmatrix}
\end{split}
\end{equation}

\noindent where each row represents an individual (i.e., candidate solution), and each column represents a dimension (i.e., variable).

\begin{figure}[ht]
    \centering
    \subfigure[The safe zone adopted by PUBG]{
    \includegraphics[width=0.45\textwidth]{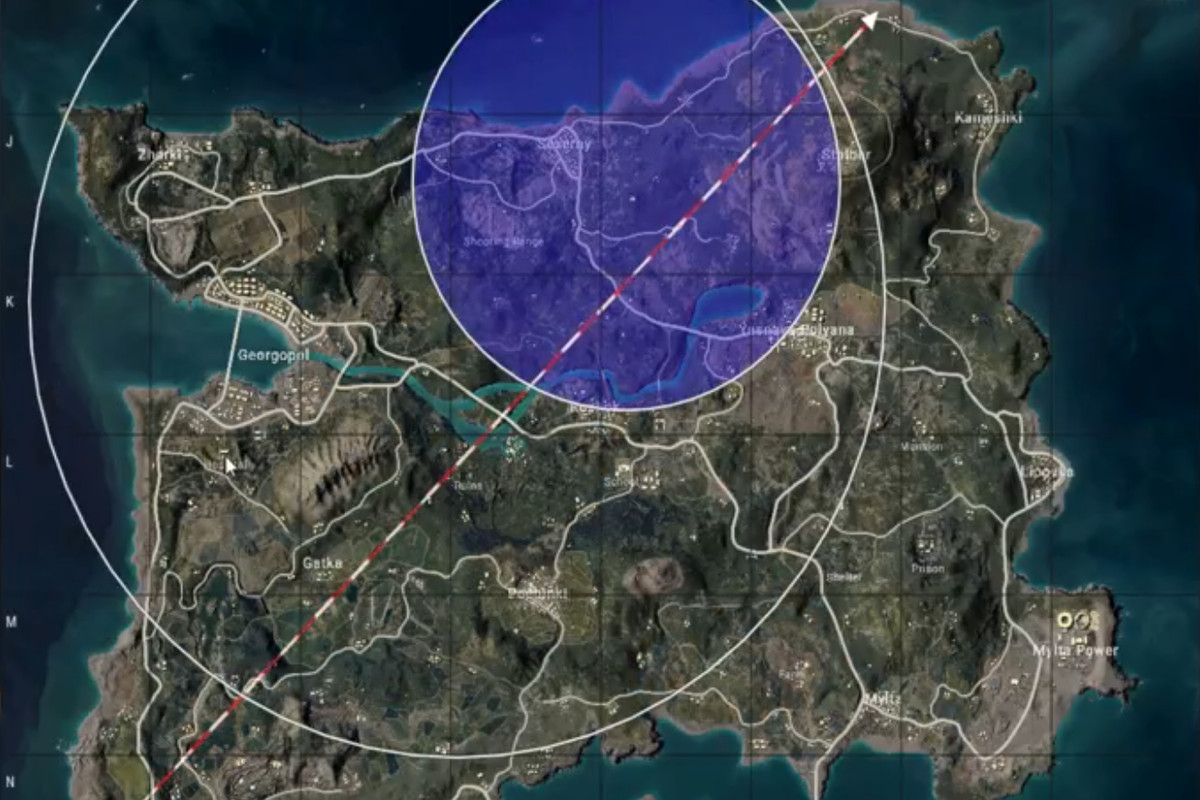}}
    \subfigure[The safe zone adopted by APEX]{
    \includegraphics[width=0.46\textwidth]{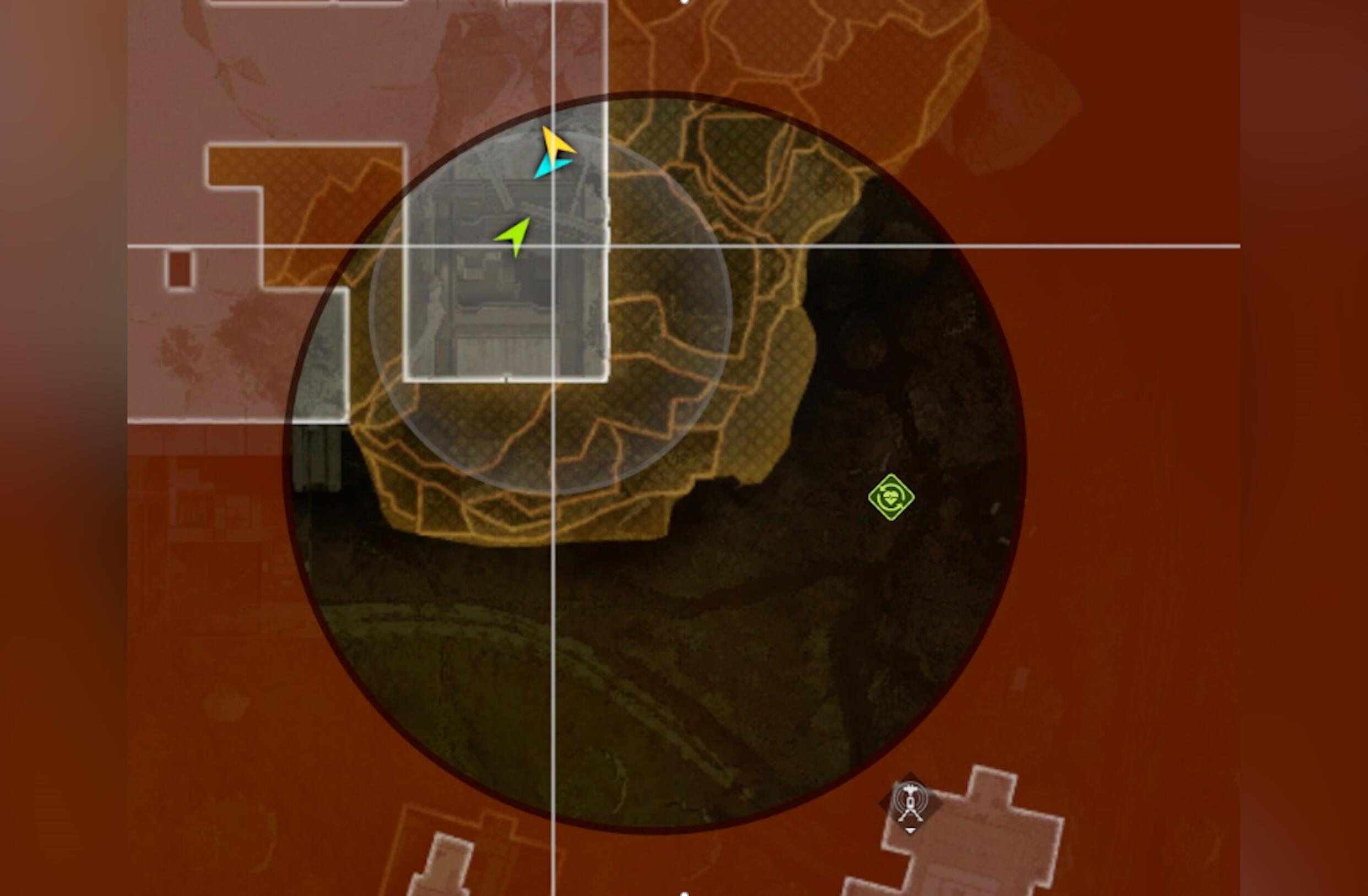}}
    \caption{Screenshots from two different games, where the entire map is divided into two different areas.\protect\footnotemark[1]}
    \label{safezone}
\end{figure}

\textbf{Movement Phase} is mainly responsible for the exploitation capability and uses the concept of the ``safe zone'' to guide individuals to converge toward potential areas. Fig. \ref{safezone} illustrates this mechanism used in two real battle royale games. Although the rules in each game may differ, the general idea is to select a part of the area as a safe area and continuously shrink the subsequent safe prefetching to encourage players to gather together as the game progresses. To simulate this mechanism, a crucial consideration is how to determine the boundaries of the safe area.

\footnotetext[1]{https://apexlegends-news.com/apex-aruaru-5/}

As an attempt, we leverage the current optimal individual, $\mathbf{x}_{best}$, and designate it as the center of the safe area. Regarding the determination of the radius covered by the safe area, we use the Euclidean distance between the best and worst individuals in the current population as the baseline.  It's important to note that the baseline distance may not necessarily be equal to the distance between the two furthest individuals in the current population. To increase randomness, this baseline distance is then multiplied by a random factor between 0.8 and 1.2, and the resulting value is utilized as the final safety radius, as shown in Eq.(\ref{eqn:2}). Thus, the entire search space dynamically divides into two distinct areas as the population converges:

\begin{equation}\label{eqn:2}
R = ( |\mathbf{x}_{best}-\mathbf{x}_{worst}|+ eps ) \times rand(0.8, 1.2)
\end{equation}

\noindent where $R$ is the safety radius, and $eps$ is a tiny non-negative number to ensure a non-zero radius for the safe area.

For individuals within the safe area, where the distance from the current best individual is less than the safety radius, it suggests that the individual is likely situated in a potential area. Here, we introduce Eq.(\ref{eqn:5}) to use the local information of the current optimal individual for the search. However, for individuals outside the safe area, meaning they are currently distant from the potential area, Eq.(\ref{eqn:4}) is employed to accelerate movement toward the current optimal area. Specifically, for each dimension $k$, one of the two methods is selected for update with equal probability. The last point that needs to be mentioned is that the elite selection is employed to update the population. This implies that an individual will only be updated if the fitness of the new individual, $\mathbf{x}_{new}$, surpasses that of the original individual, $\mathbf{x}_{i}$, otherwise, the current state will be maintained:

\begin{equation}\label{eqn:5}
\mathbf{x}_{new} = \mathbf{x}_{i}+\mathbf{x}_{best}\times \sin{(2 \times {\pi} \times rand(0, 1))}
\end{equation}

\begin{equation}\label{eqn:4}
x_{new}^{k} = \begin{cases}
x_{i}^{k} + normal(), \quad \text{if} \ rand(0, 1) < 0.5\\
x_{i}^{k} + (x_{best}^{k} - x_{i}^{k}) \times rand(0, 1), \  \text{otherwise} \\
\end{cases}
\end{equation}

\noindent where $\mathbf{x}_{new}$ and $x_{new}^{k}$ respectively represent the moved individual and its updated value in the $k$-th dimension, and $normal()$ returns a random number that adheres to the standard normal distribution.

\textbf{Battle Phase} is mainly responsible for the exploration capability and simulates diverse battle behaviors that arise when players encounter each other randomly in the game. While players may employ different strategies when facing opponents of varying skill levels, common objectives include avoiding opponent attacks to minimize damage received and inflicting maximum damage on the opponent to secure victory. To simulate the diverse behaviors of players under various psychological conditions, we have simplified the observations from real games and imposed some idealized restrictions. Here, we assume a player will not confront multiple opponents simultaneously and will, instead, randomly designate another individual as the opponent in each battle phase. Additionally, fitness information is employed to gauge the strength of an individual, aligning with factors such as player level. 

To simulate the player's behavior, we have designed two different strategies as outlined in Eqs.(\ref{eqn:6}) and (\ref{eqn:8}) to update the population. For the sake of simplicity, we have standardized the rules for selecting battle strategies for all individuals. Specifically, if the individual encounters a stronger opponent (i.e., an opponent with higher fitness), adopt Eq.(\ref{eqn:6}); otherwise, adopt Eq.(\ref{eqn:8}) to generate the new individual. Similarly, the elite selection is also applied during the battle phase:

\begin{equation}\label{eqn:6}
x_{new}^{k} = \begin{cases}
x_{i}^{k} + rand(0, 1) \times dir^{k}, \ \text{if} \ rand(0, 1) < 0.5 \\
x_{opponent}^{k} + rand(0, 1) \times dir^{k},\ \text{otherwise} \\
\end{cases}
\end{equation}

\begin{equation}\label{eqn:8}
\mathbf{x}_{new} = \mathbf{x}_{i} + \mathbf{dir} \times \cos{(2 \times {\pi} \times rand(0, 1))}
\end{equation}

\noindent where $\mathbf{dir}$ denotes the vector between the $i$-th individual and the randomly selected opponent individual $\mathbf{x}_{opponent}$. The intention is to establish the starting point of the vector at the positions of two individuals with higher fitness, thereby constructing an exclusive vector to hinder rapid convergence and prevent individuals from falling into local optimality. Additionally, the calculation of vectors is as follows:

\begin{equation}
\mathbf{dir} = \begin{cases}
x_{i}^{j} - x_{opponent}^{j}, \ \text{if} \ Fit(x_{i}) < Fit(x_{opponent}) \\
x_{opponent}^{j} - x_{i}^{j},\ \text{otherwise} \\
\end{cases}
\end{equation}

The introduction of these core components of the proposed MBGO algorithm has ended. It should be noted that the selection operation is designed into the movement phase and the battle phase, and they both involve generation updates. Algorithm \ref{alg1} finally gives a detailed overview of the proposed MBGO.

\begin{algorithm}[ht]
    \caption{The standard optimization process of the MBGO Algorithms.}
    \label{alg1}
    \begin{algorithmic}[1]
    \State Set the hyperparameters, including population size: $N$, maximum number of fitness evaluations: $t_{max}$, current number of fitness evaluations: $t=0$.
    \State Initialize the population according to Eq.(\ref{eqn:1})  and evaluate the fitness of all individuals.
     \While { $t \leq t_{max}$ }
     \State //Start the movement phase:
     \For{$i = 1$ to $N$}
     \State Determine the safety radius according to Eq.(\ref{eqn:2}).
     \If {the $i$-individual is located in a safe area}
     \State Generate $\mathbf{x}_{new}$ using Eq.(\ref{eqn:5}).
     \Else
     \State Generate $\mathbf{x}_{new}$ using Eq.(\ref{eqn:4}).
     \EndIf
     \State Evaluate the individual $\mathbf{x}_{new}$ and compare it with the $i$-individual.
     \State The winner survives and updates the $i$-individual.
     \EndFor
     \State //Start the battle phase:
     \For{$i = 1$ to $N$}
     \State Randomly select an opponent, $\mathbf{x}_{opponent}$, for the $i$-th individual.
     \If {the fitness of $\mathbf{x}_{opponent}$ is better than that of $\mathbf{x}_{i}$}
     \State Generate $\mathbf{x}_{new}$ using Eq.(\ref{eqn:6}).
     \Else
     \State Generate $\mathbf{x}_{new}$ using Eq.(\ref{eqn:8}).
     \EndIf
     \State Evaluate the individual $\mathbf{x}_{new}$ and update it by comparing it with the $i$-individual.
     \State The winner survives and updates the $i$-individual.
     \EndFor
    \EndWhile
    \State Output the optimal solution found.
\end{algorithmic}
\end{algorithm}

\section{Evaluation Experiments}
\label{sec4}
We designed a series of comparative experiments to analyze the performance of the proposed MBGO algorithm. Specifically, we selected 29 functions from the CEC2017 test suite \cite{CEC2017} and 10 functions from the CEC2020 test suite \cite{CEC2020}. These functions exhibit diverse characteristics, including multimodal, rotation, and non-separate traits, effectively covering a spectrum of optimization scenarios. Additionally, we incorporated 10 constrained engineering design problems to evaluate the scalability and potential of the proposed MBGO algorithm. As competitors, we employed 8 other EC algorithms, including three classic well-established algorithms (PSO, DE, and WOA) and 5 contemporary promising algorithms (HBA, SFO, TSA, SOA, and AO). To ensure the fairness of comparison, all algorithms were implemented in Python, and the execution environment was identical for all experiments. In addition, each algorithm was independently executed 30 times, whether applied to a benchmark function or an engineering problem.

\subsection{Performance evaluation based on two CEC suites}
\label{sec4.1}
The benchmark functions are carefully designed to compare the performance of different algorithms, with the flexibility to set different dimensions based on specific requirements. In this context, we selected widely used dimensions, specifically 10 dimensions (10-D) and 30-D, for the CEC2017 functions. Additionally, for the CEC2020 functions, we opted for higher dimensions, namely 50-D and 100-D.

The parameter configurations for the involved algorithms are as follows. The common parameters across the nine algorithms are uniformly set as follows: the population size is set to 100, and the maximum number of fitness evaluations is set to $1000 \times D$, where $D$ represents the dimensionality of the optimization problem. The parameters unique to each algorithm are aligned with mainstream settings, and their specific settings are summarized in Tab. \ref{tbl:Parameters Set}.

\begin{table}[htbp]
\scriptsize
\centering
\renewcommand\arraystretch{1.5}
\caption{The unique parameter settings for each algorithm.}
\label{tbl:Parameters Set}
\begin{tabular}{cl}
    \toprule
    Algorithm & Parameters   \\
    \midrule
        HBA & $\beta$  = 6 (Default)\\
        & $C$ = 2 (Default)\\
        
        \midrule
        PSO& $\omega$ decreases linearly from 0.9 to 0.4 (Default)\\
        & $C_1$  = 2.05 (Default)  \  $C_2$ = 2.05(Default)\\
       
        \midrule
        DE& Weighting Factor, $F$ = 0.8 (Default) \\
        & Crossover Rate, $C_r$ = 0.9 (Default) \\

        \midrule
        WOA& $a$ variable decreases linearly from 2 to 0 (Default)\\
        & $a2$ linearly decreases from 1 to 2 (Default)\\

        \midrule
        TSA& Parameter $P_{min}$  = 1 (Default)\\
        & Parameter $P_{max}$  = 4 (Default)\\

        \midrule
        SFO& Parameter $A$  = 4 (Default)\\
        & Parameter $\epsilon$  = 0.001 (Default)\\
        
        \midrule
        SOA& Parameter $A$ is random from 2 to 0 (Default)\\
        &$f_c$  = 2 (Default)\\

        \midrule
        AO& $G1 = 2\times rand$ - 1 (Default)\\
        &$G2$  decreasing values from 2 to 0 (Default)\\

        \midrule
        MBGO&$\alpha$ is random from 0.8 to 1.2 (Default)\\
    \bottomrule
\end{tabular}
\end{table}

\begin{figure*}
    \centering
    \subfigure[F1]{
    \includegraphics[width=0.2\textwidth]{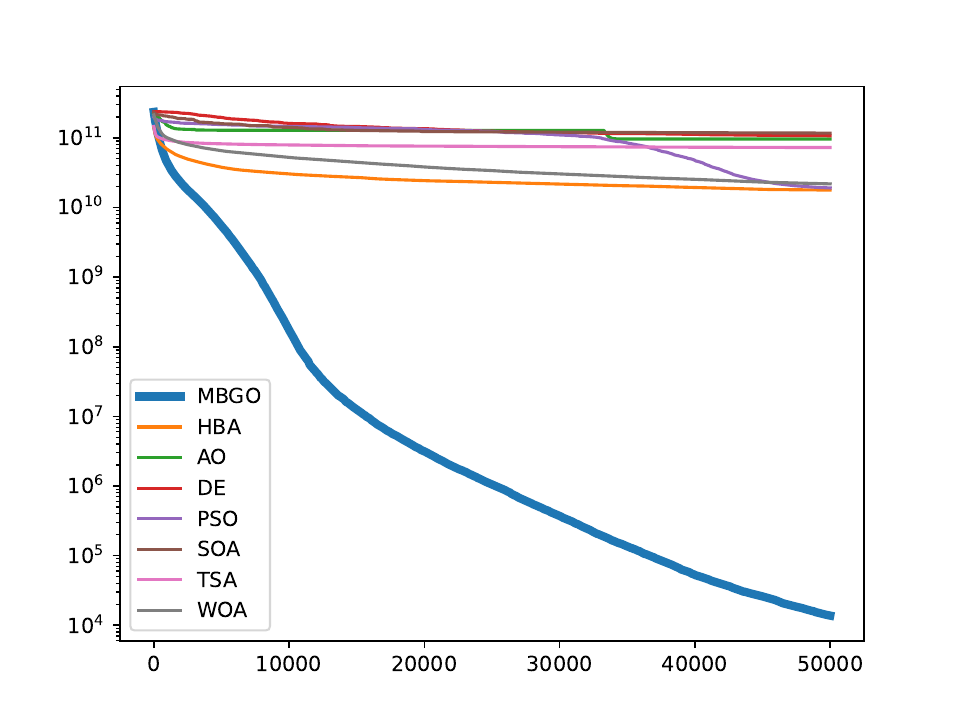}}
    \hspace{-6.2mm}
    \subfigure[F2]{
    \includegraphics[width=0.2\textwidth]{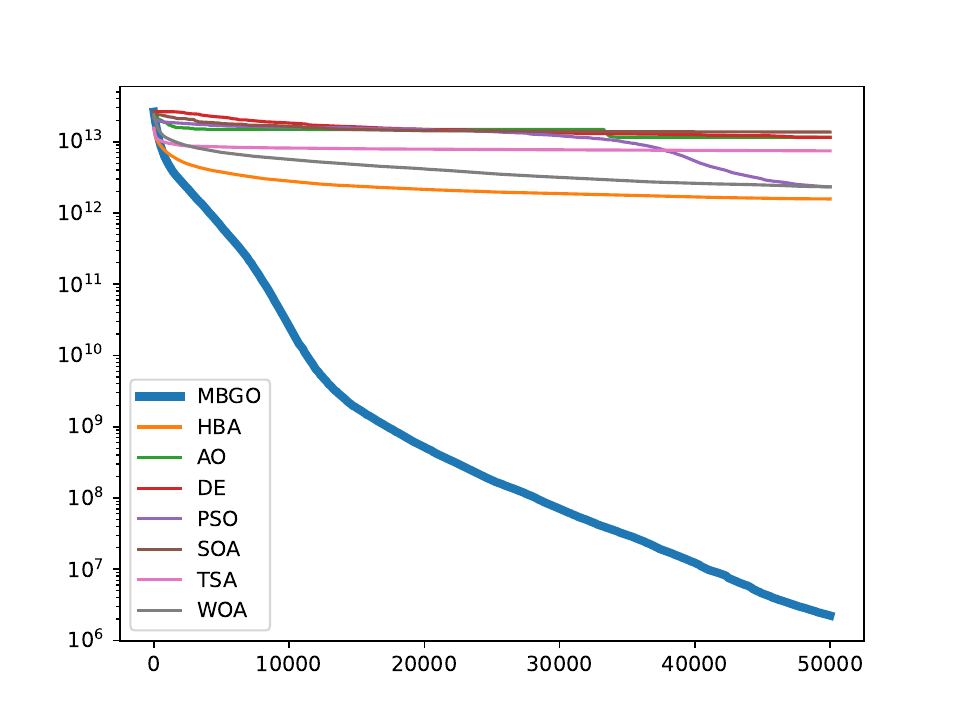}}
    \hspace{-6.2mm}
    \subfigure[F3]{
    \includegraphics[width=0.2\textwidth]{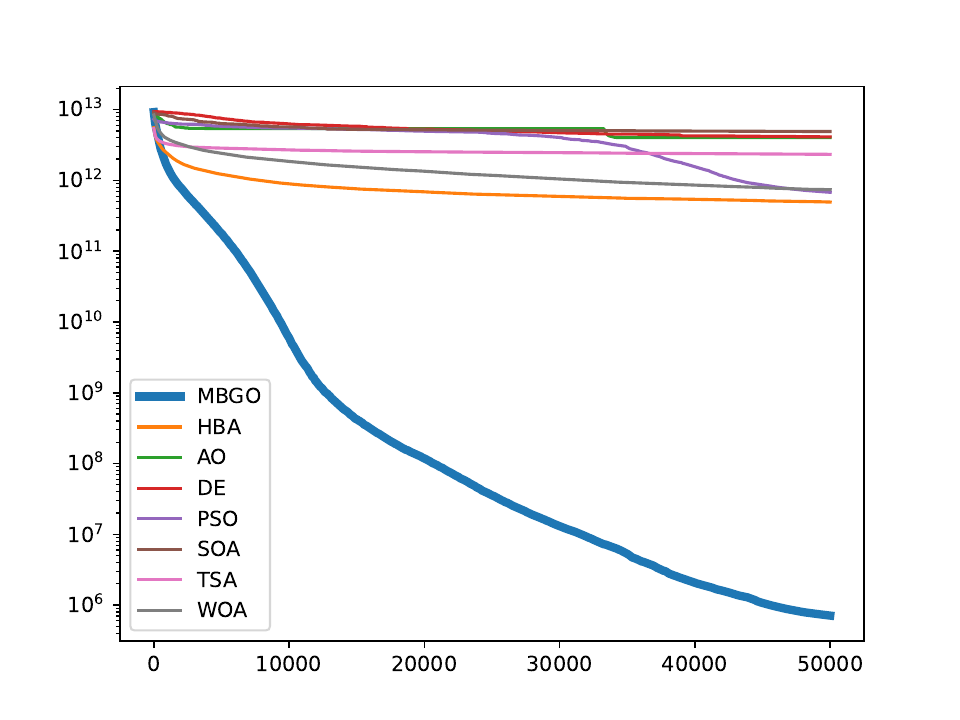}}
    \hspace{-6.2mm}
    \subfigure[F4]{
    \includegraphics[width=0.2\textwidth]{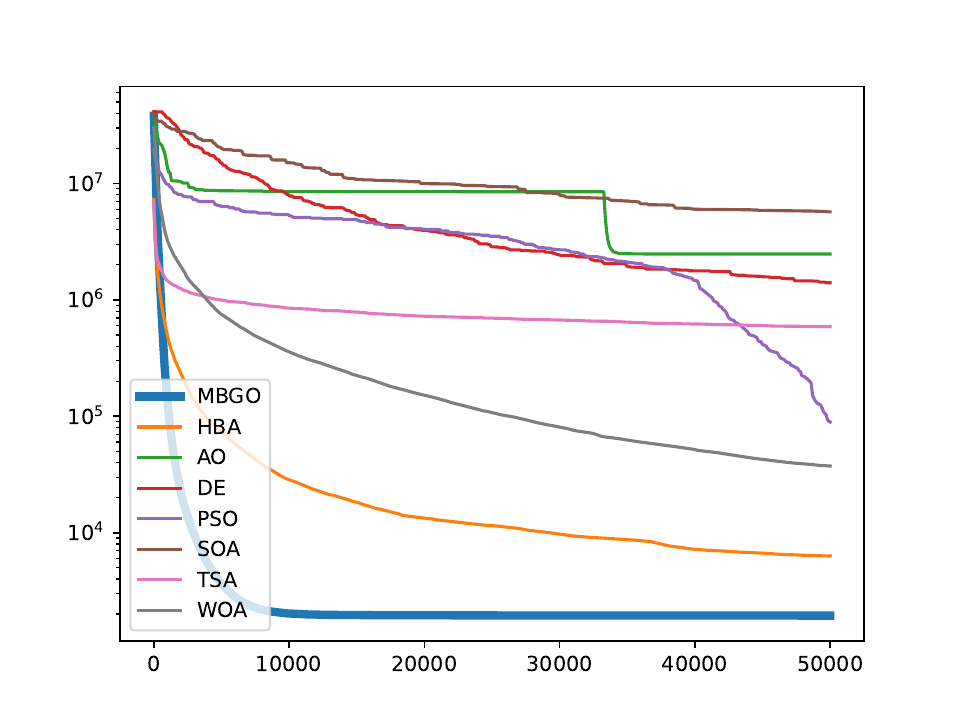}}
    \hspace{-6.2mm}
        \subfigure[F5]{
    \includegraphics[width=0.2\textwidth]{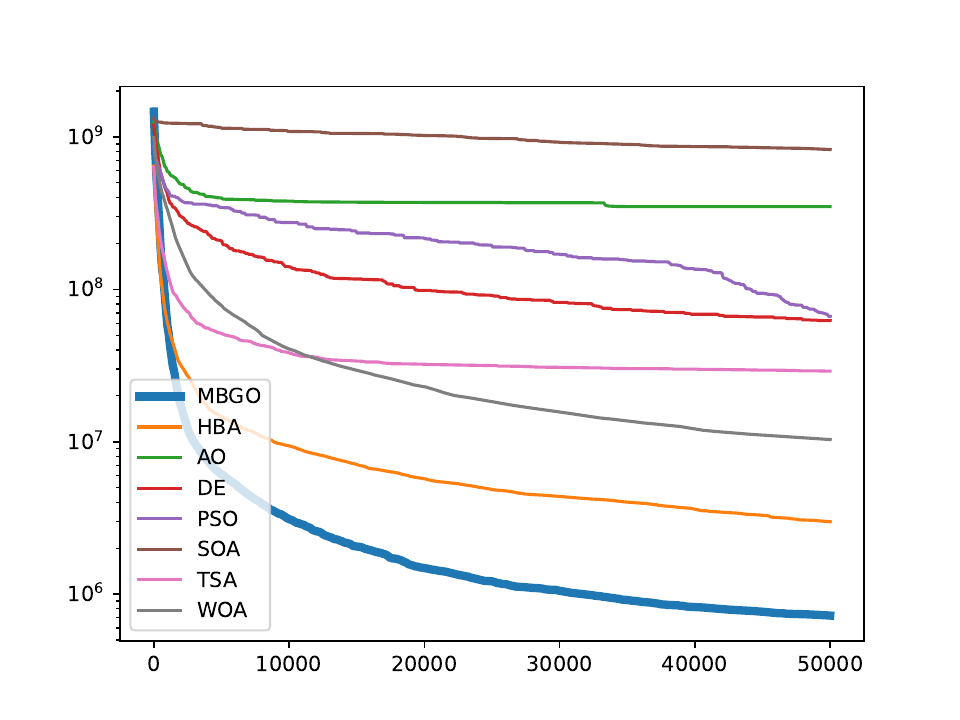}}
    \hspace{-6.2mm}
        \subfigure[F6]{
    \includegraphics[width=0.2\textwidth]{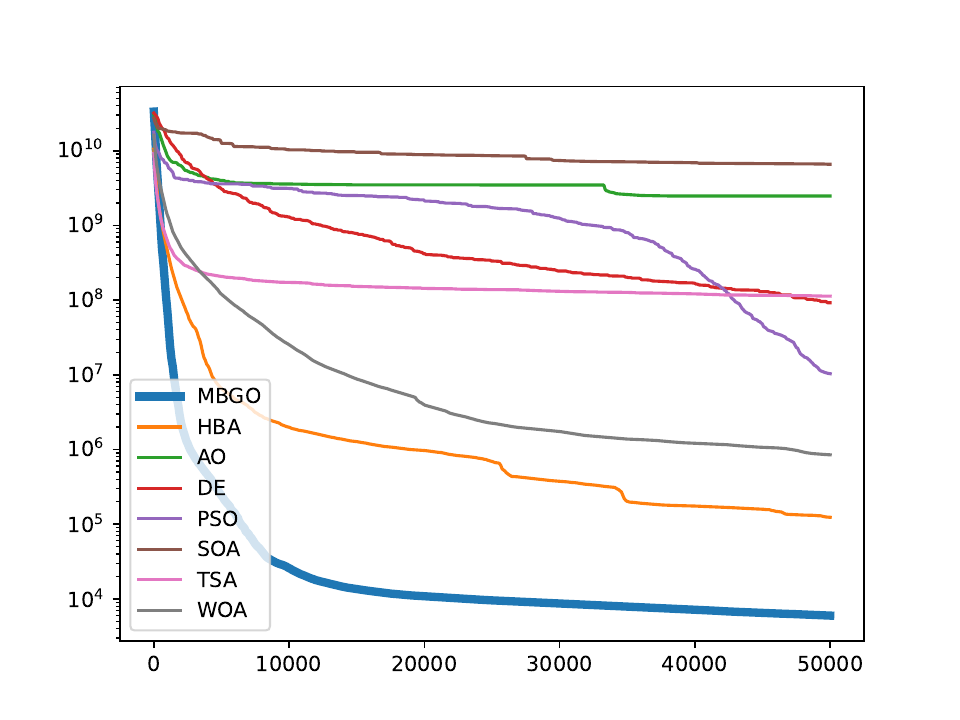}}
    \hspace{-6.2mm}
        \subfigure[F7]{
    \includegraphics[width=0.2\textwidth]{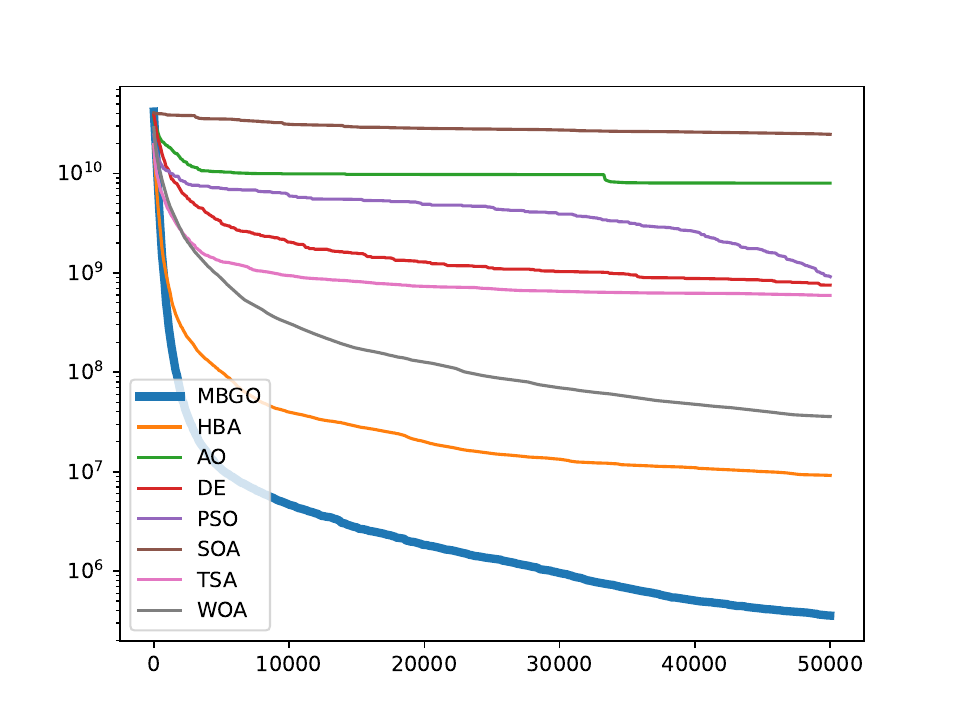}}
    \hspace{-6.2mm}
        \subfigure[F8]{
    \includegraphics[width=0.2\textwidth]{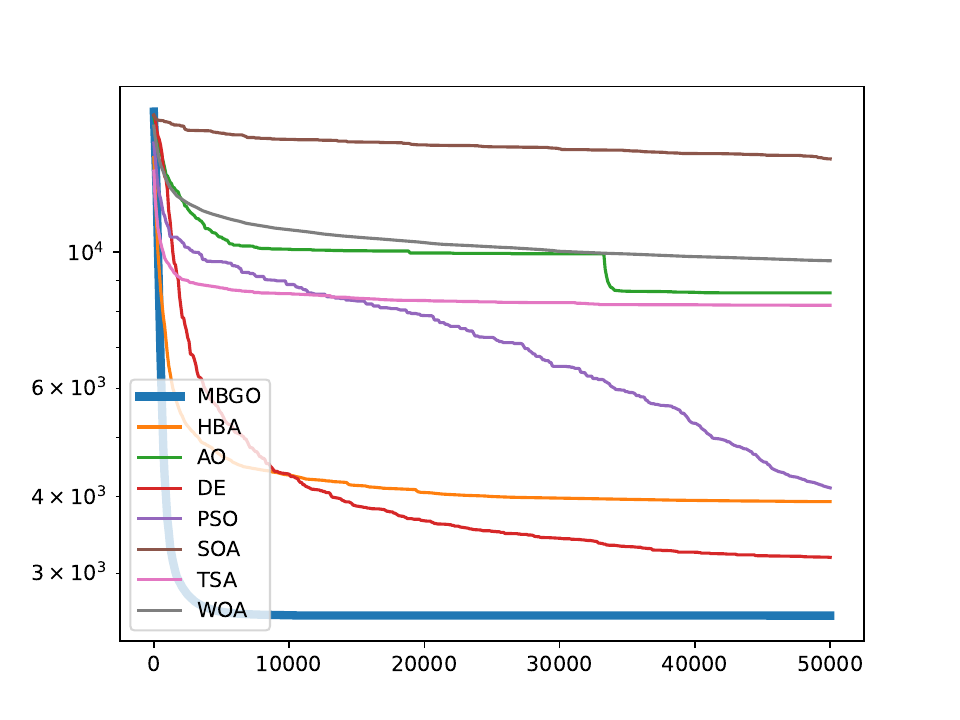}}
    \hspace{-6.2mm}
        \subfigure[F9]{
    \includegraphics[width=0.2\textwidth]{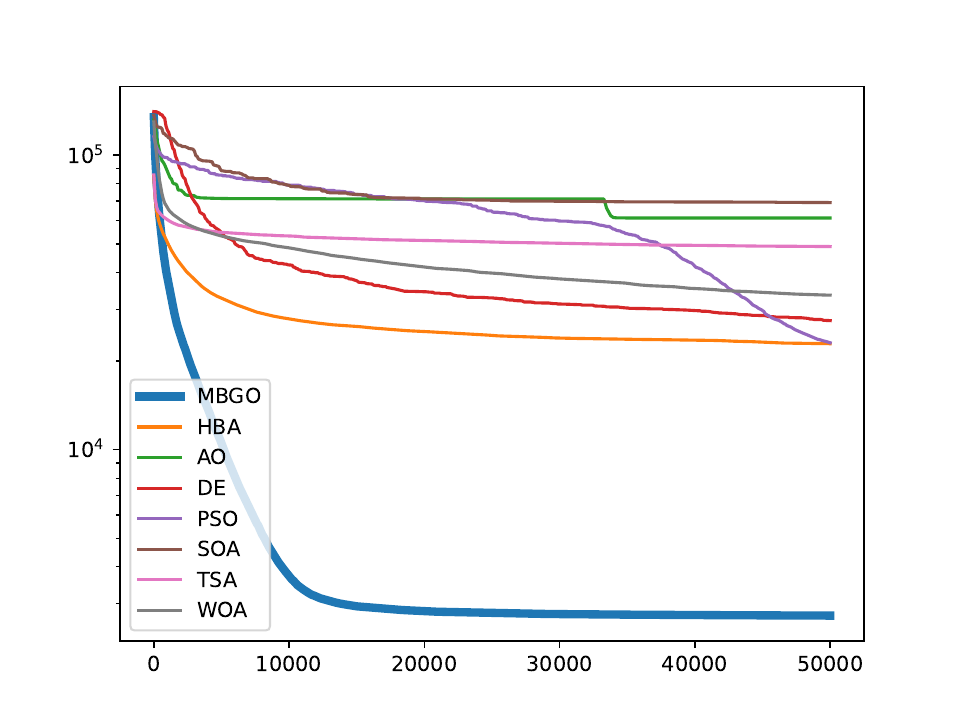}}
    \hspace{-6.2mm}
        \subfigure[F10]{
    \includegraphics[width=0.2\textwidth]{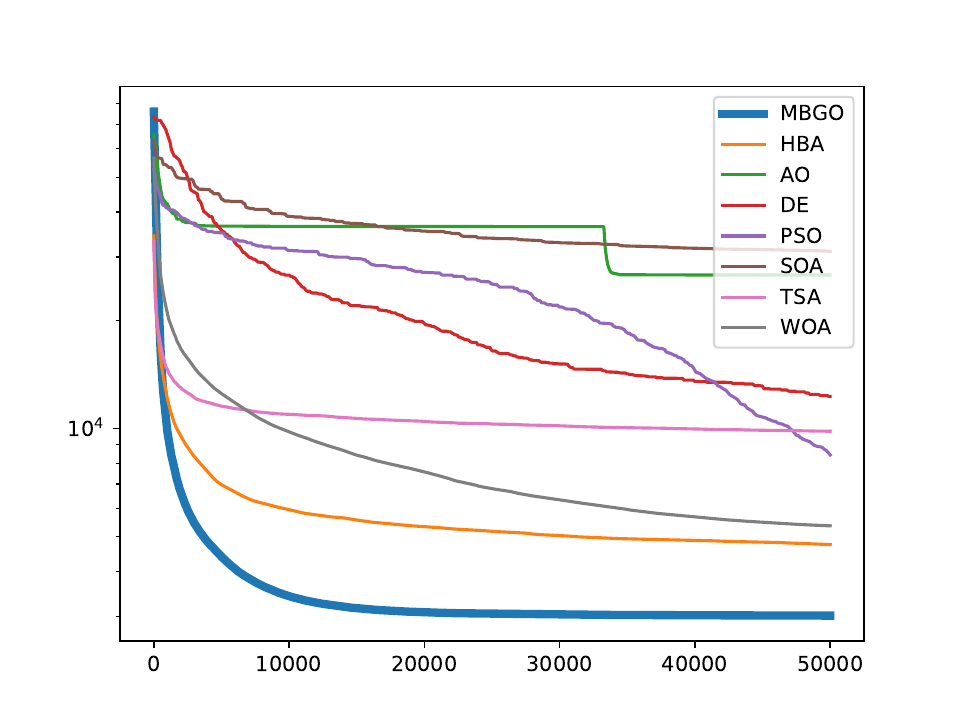}}
    \caption{The average convergence curves of all competitor algorithms on 50-D CEC2020 functions.The horizontal coordinate represents the total number of evolutions of the current individual and the vertical coordinate represents the fitness value of the current optimal solution.}
    \label{CEC2020_50D}
\end{figure*}

\begin{figure*}
    \centering
    \subfigure[F1]{
    \includegraphics[width=0.2\textwidth]{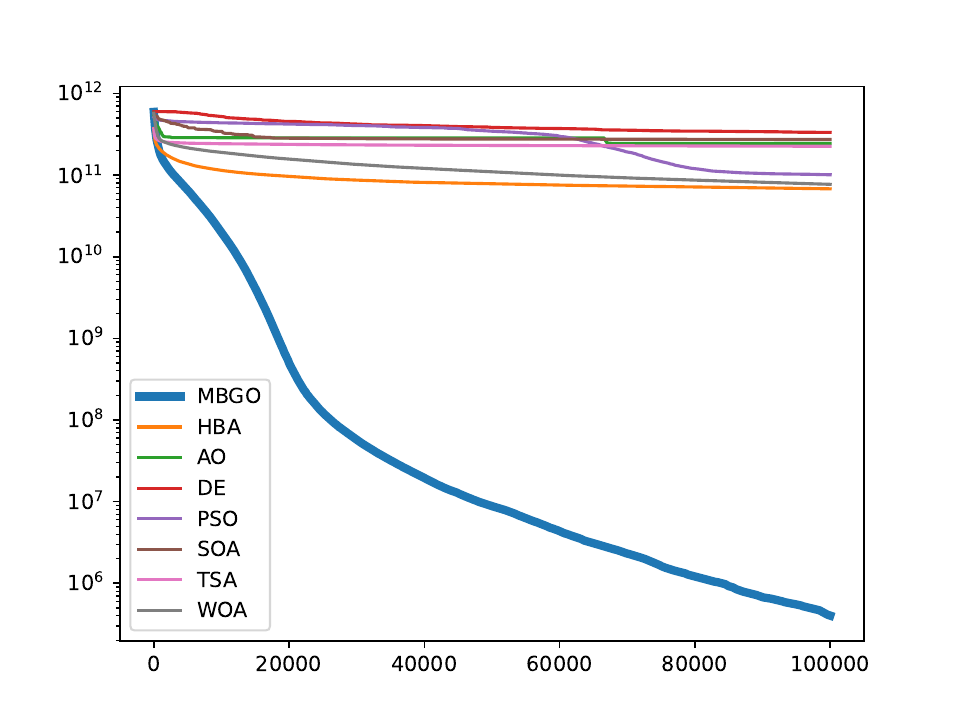}}
        \hspace{-6.2mm}
    \subfigure[F2]{
    \includegraphics[width=0.2\textwidth]{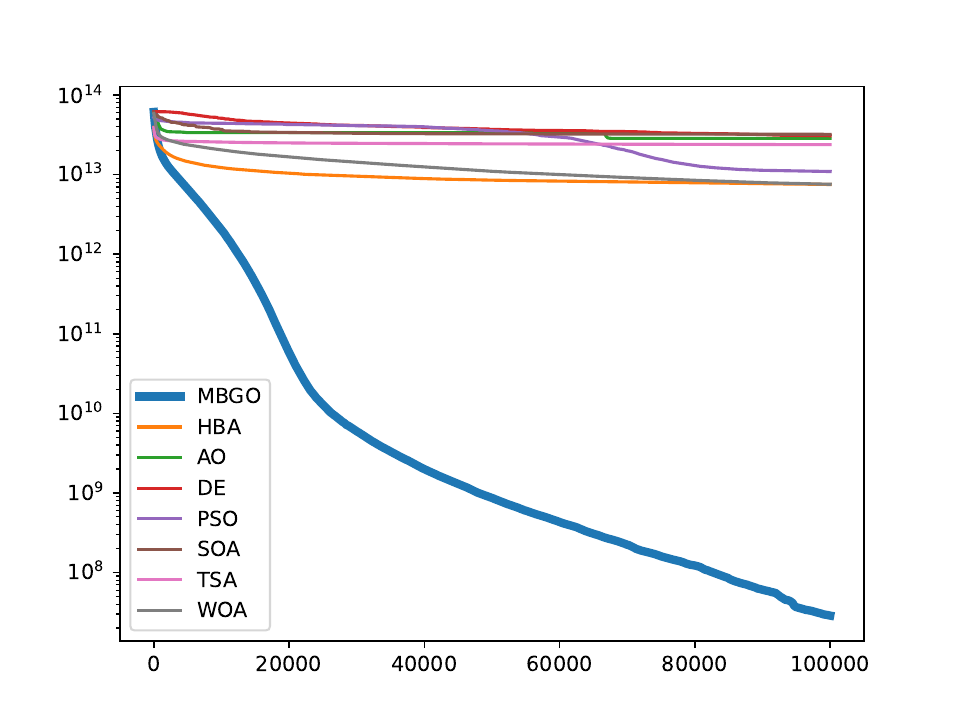}}
        \hspace{-6.2mm}
    \subfigure[F3]{
    \includegraphics[width=0.2\textwidth]{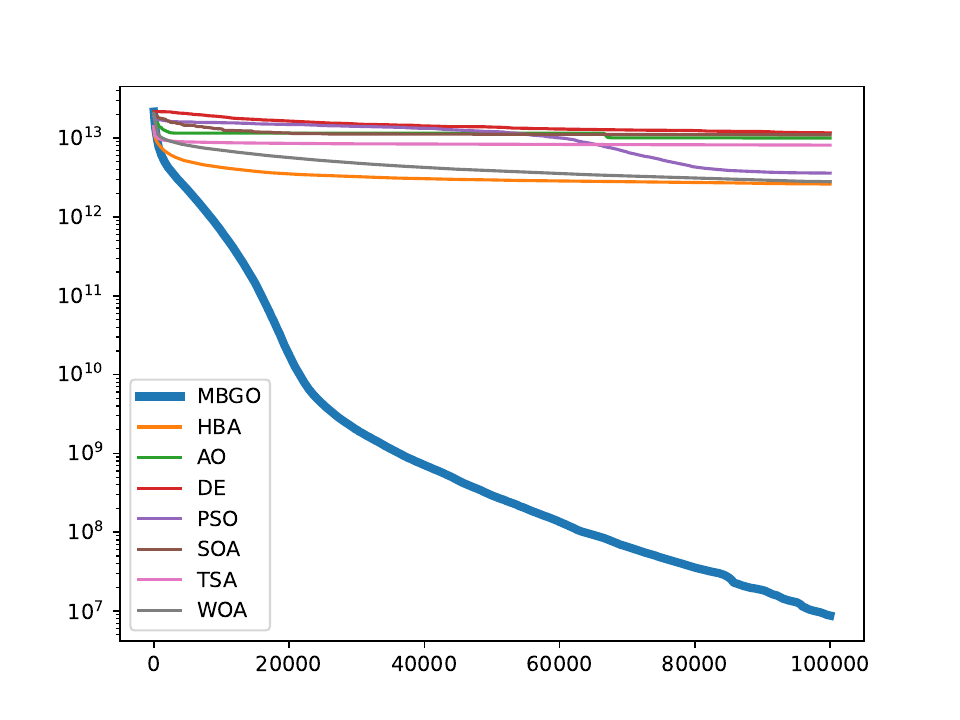}}
        \hspace{-6.2mm}
    \subfigure[F4]{
    \includegraphics[width=0.2\textwidth]{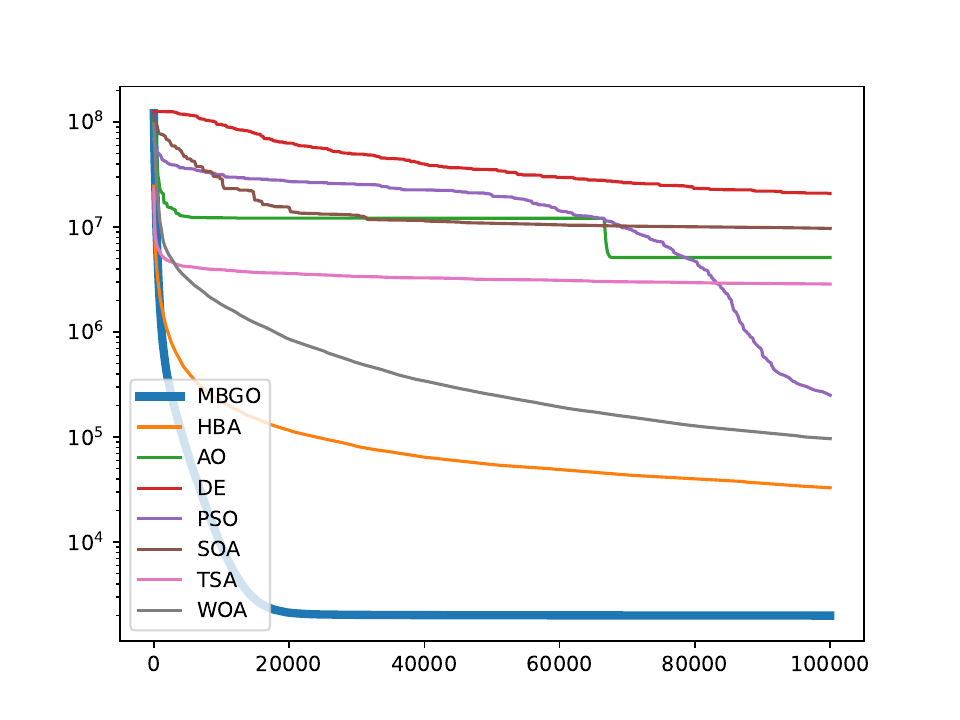}}
        \hspace{-6.2mm}
        \subfigure[F5]{
    \includegraphics[width=0.2\textwidth]{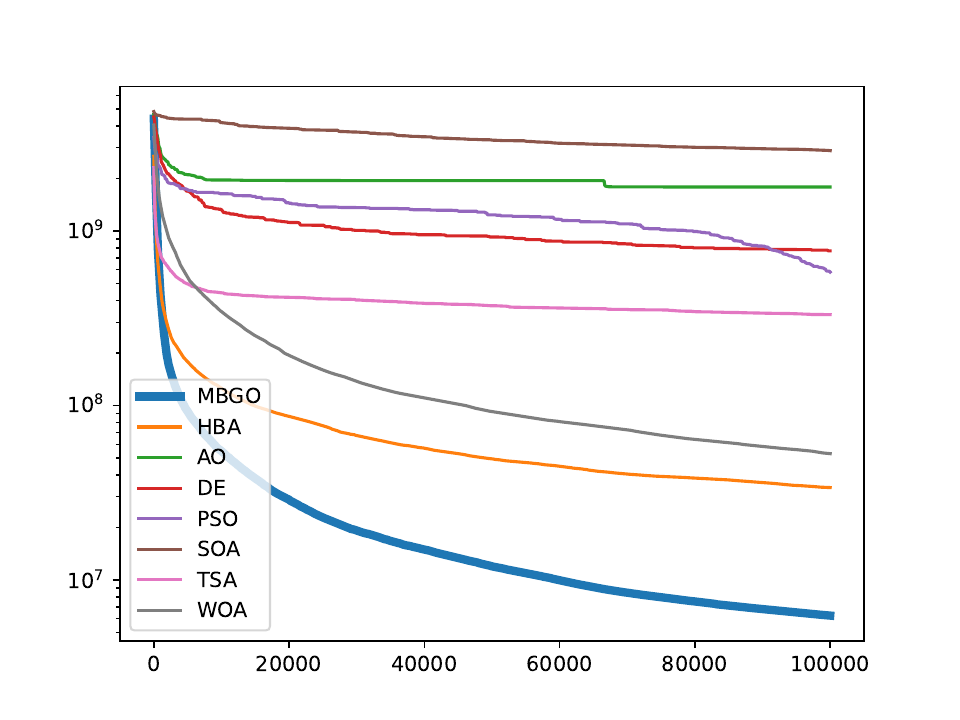}}
        \hspace{-6.2mm}
        \subfigure[F6]{
    \includegraphics[width=0.2\textwidth]{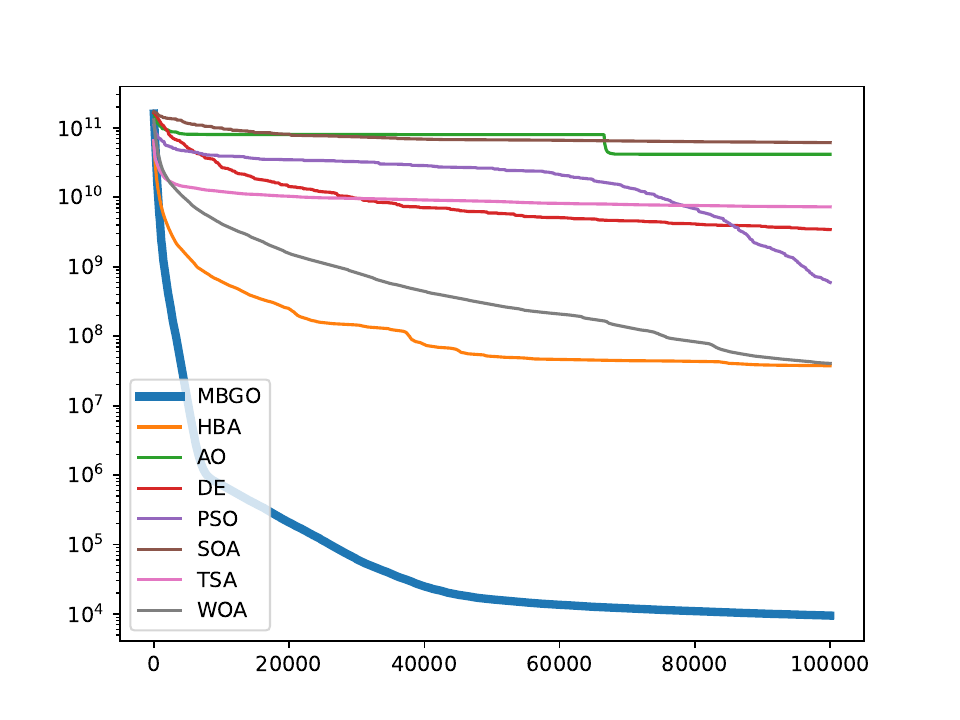}}
        \hspace{-6.2mm}
        \subfigure[F7]{
    \includegraphics[width=0.2\textwidth]{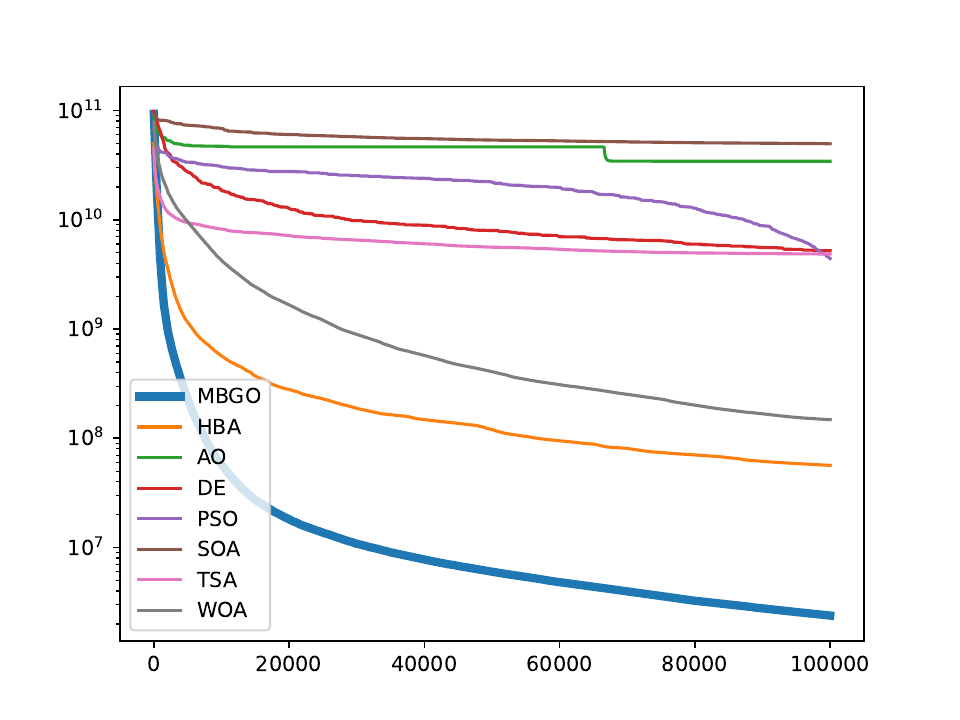}}
        \hspace{-6.2mm}
        \subfigure[F8]{
    \includegraphics[width=0.2\textwidth]{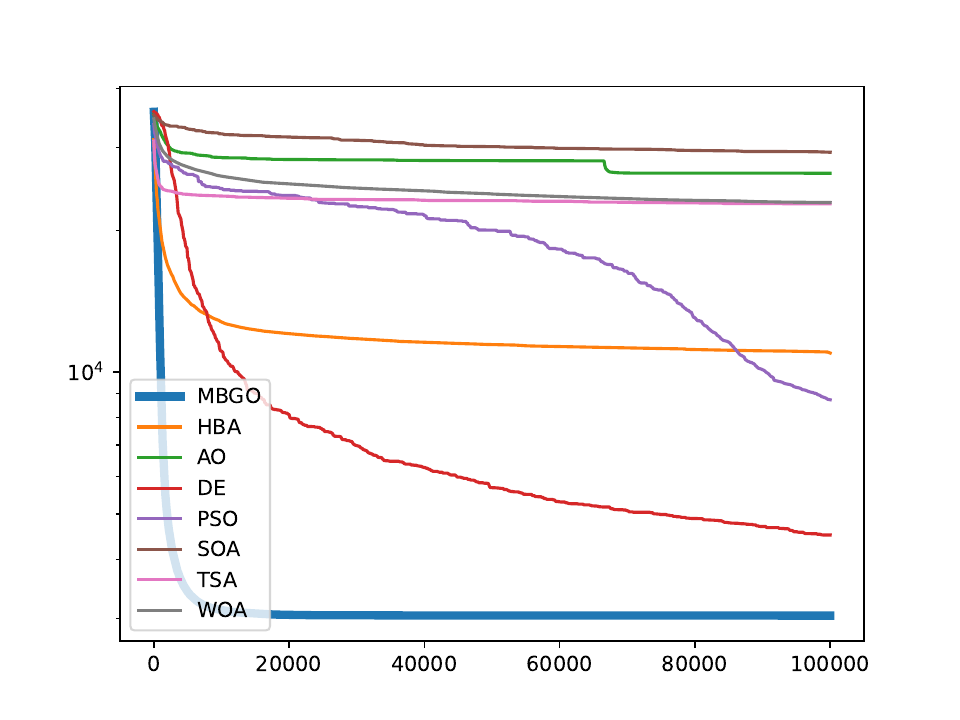}}
        \hspace{-6.2mm}
        \subfigure[F9]{
    \includegraphics[width=0.2\textwidth]{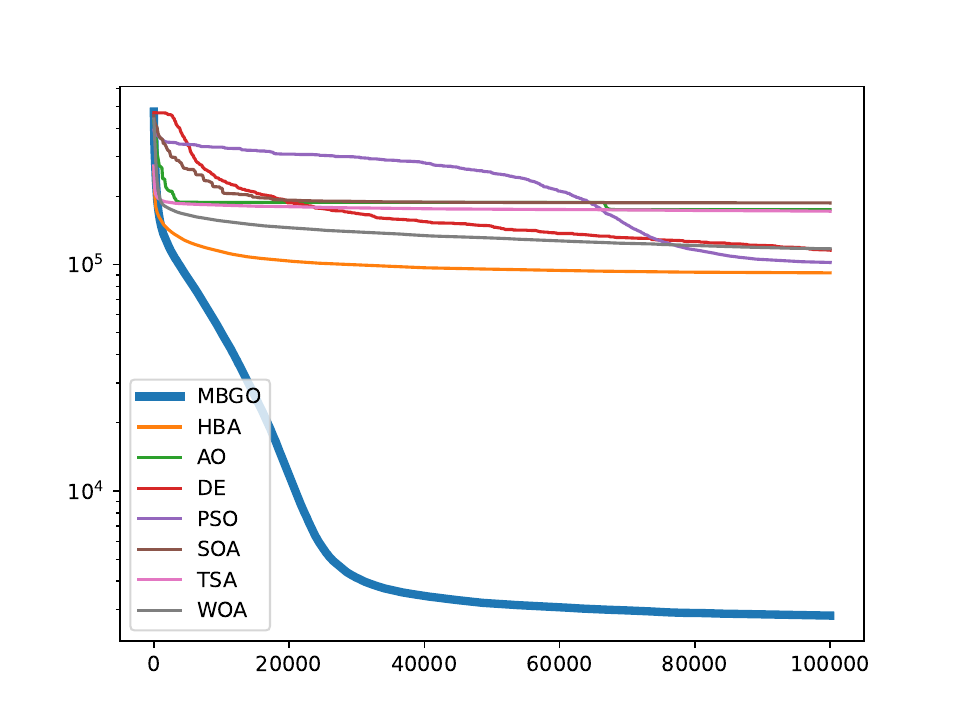}}
        \hspace{-6.2mm}
        \subfigure[F10]{
    \includegraphics[width=0.2\textwidth]{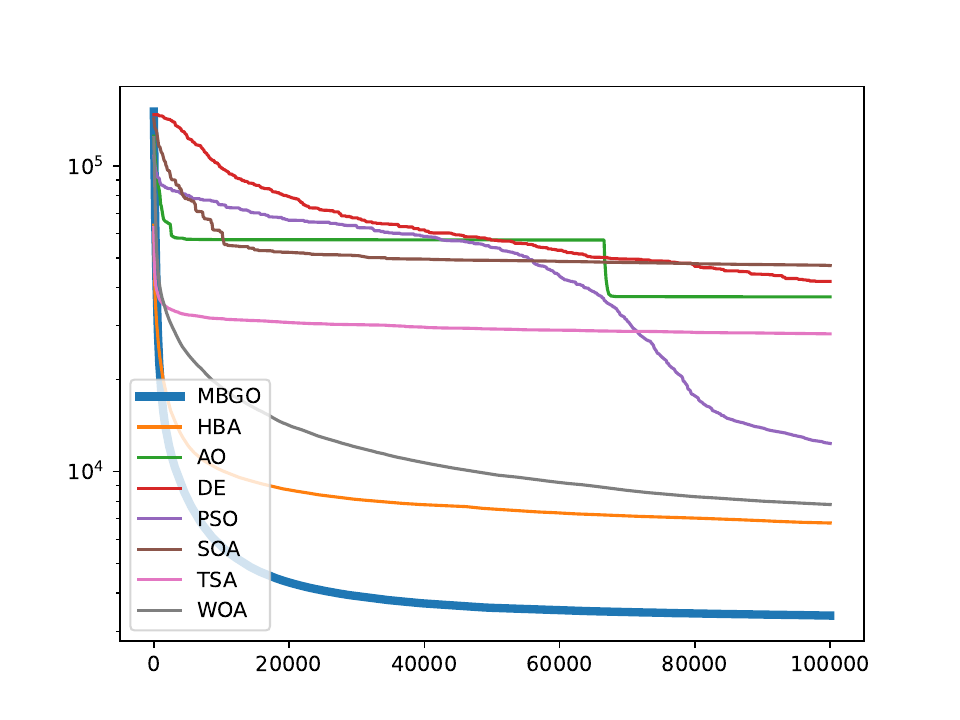}}
    \caption{The average convergence curves of all competitor algorithms on 100-D CEC2020 functions. The horizontal coordinate and the vertical coordinate are the same as Fig. \ref{CEC2020_50D}.}
    \label{CEC2020_100D}
\end{figure*}

To investigate whether there are significant differences between the proposed algorithm and other algorithms, we conducted the U-test and Holm's multiple comparisons at the termination of the algorithm, that is, at the maximum number of fitness evaluations. Additionally, we present the average and standard deviation of the optimal solution found in the final 30 trial runs for all algorithms. Tab. \ref{tbl:CEC2017_10D} and Tab. \ref{tbl:CEC2017_30D} summarize the results of 10-D and 30-D on CEC2017 test suites, and Tab. \ref{tbl:CEC2020_50D} and Tab. \ref{tbl:CEC2020_100D} summarize the results of 50-D, and 100-D on CEC2020 test suites, respectively. Besides, we present the average convergence curves of all algorithms in both 50-D and 100-D for all CEC2020 functions, as illustrated in Figs. \ref{CEC2020_50D} and \ref{CEC2020_100D}.

\subsection{Performance evaluation based on real-world problems}
\label{sec4.2}

Not limited to human-designed benchmark functions, we employed 10 real industrial design problems to evaluate the performance of the proposed MBGO algorithm. These real-world problems are frequently characterized by robust constraints, posing challenges that enhance the complexity of the search for optimal solutions. Since none of the algorithms inherently addresses constraints, here a unified penalty function is employed in this context, which serves to impose an infinite penalty on the fitness of individuals that violate constraints, reflecting the most unfavorable outcomes. The details of the ten engineering design problems are outlined in Tab. \ref{tbl:real-world engineers}.

\begin{table}[htbp]
	\scriptsize
	\centering
	\renewcommand\arraystretch{1.5}
	\caption{Eight Realistic Engineering Optimization Problems Summary.}
	\label{tbl:real-world engineers}
	\resizebox{\columnwidth}{!}{
	\begin{tabular}{cl}
		\toprule
		Pro. & Description   \\
		\midrule
            WBP& Design cost minimization optimization problem for welded beams. \\
            \midrule
            PVP & The problem of minimizing the total cost of a cylindrical vessel capped by a hemispherical head.\\
            \midrule
            TBTD & Minimize three-bar structure weight subject to supporting a total load P acting vertically downwards.\\
            \midrule
            GTD & Unconstrained discrete design optimization problem.  \\
            \midrule
            CBD & Minimize a cantilever beam's weight.  \\
            \midrule
            IBD & Minimizes the vertical deflection of a beam.  \\
            \midrule
            TCD & Optimize the cost of the column in terms of being able to resist torsion, bending and shear. \\
            \midrule
            PLD & Piston rod cost optimization problem.  \\
            \midrule
            CBHD &  Corrugated Groove Head Cost Optimization Problem. \\
            \midrule
            RCB &  Cost minimization problem for reinforced concrete beams. \\
		\bottomrule
	\end{tabular}
        }
\end{table}

The parameter configuration for all algorithms remains identical to the previous experiment, with the sole exception being the maximum number of fitness evaluations. Since the dimensions of these real-world problems vary, we standardized the maximum number of fitness evaluations for all problems to 20,000. To analyze the performance between algorithms, we once again employ the U-test and Holm's multiple comparisons to determine whether there is a significant difference at the end of all algorithms. Tab. \ref{tbl:real-world problems results} summarizes the average and standard deviation of optimal solutions discovered by these algorithms, along with the outcomes of statistical testing.

\section{Discussion}
\label{sec5}

\subsection{Computational complexity of the proposed MBGO}
For the sake of simplicity, we continue to use the symbols introduced in Sec. \ref{sec3} and define the maximum number of fitness evaluations as $T$. Therefore, the complexity of the core operations of the MBGO algorithm is as follows:

\begin{itemize}
\item initialization: $O(N \times D)$.
\item the selection for current best individual: $O(N)$.
\item the selection for current worst individual: $O(N)$.
\item the movement phase: $O(N \times D)$.
\item the battle phase: $O(N \times D)$.
\end{itemize}

In summary, the computational complexity of the MBGO algorithm can be calculated by Eq.(\ref{eqn:9}).
\begin{equation}\label{eqn:9}
\begin{split}
&O(N\times D+T \times (N+N+N \times D+N \times D))\\
=&O(N \times D+T \times N(2+D))\\
=&O(N \times D+T \times N \times D))\\
=&O(T \times N \times D)
\end{split}
\end{equation}

\subsection{Benefits from the core components}
The proposed MBGO primarily consists of three operations: initialization, the movement phase, and the battle phase. Excluding the initialization operation applied in all comparison algorithms, the remaining two operations significantly contribute to performance improvement. This fundamental distinction is the key reason for the superior performance of the proposed MBGO algorithm compared to others. Here, we conduct an in-depth analysis of the two operations and enumerate the benefits they bring.

The movement phase uses the Euclidean distance between the best individual and the worst individual to divide the entire search space into potential (safe) areas and non-potential (unsafe) areas. Individuals in different areas employ varied strategies to update their current positions, effectively reducing the risk of rapid convergence that may confine the population to local areas, thereby preventing premature maturation. Moreover, as the population converges, individuals gradually converge toward the global area, resulting in the gradual reduction of the distance between the best individuals and the worst individuals. This effectively simulates the widespread adoption of ``safe zones'' in the game, which gradually shrinks to a smaller area. Furthermore, the optimal individual of each generation is designated as the center of the safe area, offering effective guidance to all other individuals to reduce random searches and accelerate overall processes. We thus can say that the movement phase guarantees the convergence speed of the proposed MBGO algorithm.

The battle phase simulates various player behaviors in response to opponents, aiming to increase the diversity of the population. Each individual randomly selects an opponent and employs different update strategies based on the opponent's fitness, fostering information exchange among individuals. Furthermore, the adversaries individuals encounter in different generations undergo constant changes, fundamentally heightening the diversity of the population and facilitating the generation of diverse offspring individuals. Additionally, we intentionally use vector information from better individuals to worse individuals to encourage the exploration of unknown potential areas and prevent premature convergence. This approach is inspired by the concept of simulated annealing. However, it is essential to clarify that the pursuit of diversity does not imply the unconditional acceptance of inferior individuals. In this context, the elite selection is implemented to ensure that population diversity is maintained without weakening the population.

Excessive emphasis on either convergence speed or diversity can pose challenges for the population to converge to the global optimum. Striking the right balance between these factors is crucial for the effective optimization of the algorithm. The collaborative synergy between the two stages enables the proposed MBGO algorithm to achieve a harmonious balance between exploration and exploitation, distinguishing itself prominently among all competing algorithms when facing various optimization problems. In addition, the rational use of individual distribution information and fitness landscape information also helps the MBGO algorithm select appropriate search strategies, ensuring sustained high search performance.

\subsection{Analysis of statistical results}
The results of statistical tests demonstrate that the proposed MBGO algorithm consistently outperforms others, showcasing superior performance across both manually designed functions and real engineering problems. In the case of the CEC2017 test suite, the MBGO algorithm demonstrates superiority over all but HBA in low-dimensional problems. However, when considering the overall performance, the MBGO algorithm stands out as excellent. Notably, it exhibits a growing advantage as dimensionality increases, leading to the achievement of the best performance on almost all problems. Moreover, in the evaluation across all functions of CEC2020, the MBGO algorithm overwhelmingly outperforms all comparison algorithms. The convergence curves depicted in Fig. \ref{CEC2020_50D} and  Fig. \ref{CEC2020_100D} further show that the proposed MBGO algorithm not only achieves high convergence accuracy but also exhibits the characteristic of rapid convergence speed. Through the analysis and summary of all function characteristics, it becomes evident that the MBGO algorithm performs better on complex problems with high dimensions and multiple peaks.

Despite not meticulously designing the constraint processing module and opting for the widely used penalty function, experimental results indicate that the proposed MBGO algorithm consistently discovers satisfactory feasible solutions when compared with other classic EC algorithms. These strong constraints, result in a substantial number of infeasible solutions and complicate the effectiveness of the search process. Overall, despite the challenges posed by strong constraints, the MBGO algorithm remains competitive and ranks among the best. This observation underscores the potential applicability of the MBGO algorithm to constrained optimization problems.

\subsection{Potential analysis and open topics}
While we have highlighted the strengths of the MBGO algorithm, it is undeniable that there is still considerable room for improvement. In this context, we present some open topics with the hope of sparking new inspiration for those who follow in our footsteps.

1. The interactive switching between the two stages is a key factor in the success of the MBGO algorithm. Exploring how to make informed and context-specific switches based on the characteristics of the optimization problem is a worthwhile research topic. Instead of treating the two stages equally, tailoring the switching mechanism to the specific nuances of the problem could enhance the MBGO algorithm's adaptability and performance.

2. While elite selection ensures that the population converges toward the optimal area, it introduces a new challenge: the direct discarding of inferior offspring individuals without any utilization. This operation can lead to a waste of evaluation resources, particularly for expensive problems. Thus, a valuable research topic involves the reuse of these discarded individuals, aiming to further enhance search performance.

3. Real games often feature intricate and engaging gameplay mechanics. In this context, we simplify these complex mechanisms to design optimized game mechanics. Thus, introducing these diverse strategies into subsequent improved versions could be an interesting research topic. Additionally, modeling the behavior of game players, aiming to replicate battle behavior more realistically, is a subject worthy of further study.

4. The hyperparameters of optimization algorithms frequently exert a significant influence on performance. Here, we simplify the process by unifying and fixing these parameters. However, it's important to note that this strategy may constrain performance to a certain extent, especially when applied to problems with diverse characteristics. Thus, exploring adaptive or problem-specific tuning of hyperparameters could be a valuable avenue for improvement in future iterations of the algorithm.

Certainly, the scope of research is not limited to the topics listed. There are numerous areas worthy of study, such as extending the MBGO algorithm to different optimization problems, including multi-objective optimization, dynamic optimization, and multi-task optimization, among others. Exploring these avenues can contribute to the MBGO algorithm's versatility and applicability across a broader range of problem domains.

\section{Conclusion}
\label{sec6}
We present a novel population-based heuristic algorithm named the multiplayer battle game-inspired optimizer (MBGO), inspired by the common mechanics of multiplayer battle royale and player behaviors in various scenarios. The proposed MBGO algorithm strategically balances exploration and exploitation by alternating between the movement phase and the battle phase in turns, enhancing its adaptability to optimization problems with diverse characteristics. The experimental results confirm the effectiveness and competitiveness of the new MBGO algorithm, whether applied to solving human-designed functions or real industrial problems.

In our forthcoming work, we intend to introduce more efficient search strategies to further enhance the performance of the proposed MBGO algorithm. We also aim to broaden the scope to include various optimization problems, such as multi-objective optimization, dynamic optimization, and constrained optimization. Additionally, we plan to conduct a theoretical analysis of the effective mechanism of the MBGO algorithm and work towards establishing feedback between the optimization problem and the algorithm to reduce overall optimization costs.

\bibliographystyle{unsrt}
\bibliography{template}

\begin{sidewaystable*}[!ht]
	\scriptsize
	\centering
	\renewcommand\arraystretch{1.2}
	\caption{The mean and standard deviation of the optimal solutions obtained from 30 trial runs for all algorithms are reported for the 10-D CEC2017 function functions. In this context, symbols $+$ and $-$ respectively signify that the proposed algorithm outperforms or underperforms the comparison algorithm. The symbol $=$ denotes no significant difference between the two. Furthermore, for each function, the optimal solutions identified among all algorithms are highlighted in bold.The last row of the table represents the ``number of functions that outperform the comparison algorithm / number of functions that are significantly different / number of functions that are lower than the comparison algorithm''.}
	\label{tbl:CEC2017_10D}
	\resizebox{\linewidth}{!}{
		\begin{tabular}{cccccccccccccccccccccc}
		  \toprule
			\multirow{3}{*}{Func} & \multicolumn{2}{c}{DE} & \multicolumn{2}{c}{PSO} & \multicolumn{2}{c}{AO} & \multicolumn{2}{c}{SOA} &
            \multicolumn{2}{c}{SFO} & 
            \multicolumn{2}{c}{WOA} & 
            \multicolumn{2}{c}{HBA} & 
            \multicolumn{2}{c}{TSA} & 
            \multicolumn{2}{c}{MBGO}
   \\ \\
			\cmidrule(r){2-3} \cmidrule(r){4-5} \cmidrule(r){6-7} \cmidrule(r){8-9} \cmidrule(r){10-11} \cmidrule(r){12-13} \cmidrule(r){14-15} \cmidrule(r){16-17} \cmidrule(r){18-19} 
                & mean & std & mean & std & mean & std & mean & std
                & mean & std & mean & std
                & mean & std & mean & std
                & mean & std
                \\\bottomrule

            \midrule
            F1	&7.6447E+08	+	&1.6147E+08	&3.6899E+08	+	&5.8412E+08	&4.2405E+09	+	&2.3757E+09	&1.2531E+10	+	&3.6743E+09	&2.4043E+09	+	&8.8332E+08	&6.1455E+07	+	&9.4622E+07	&1.4774E+06	+	&9.3662E+05	&6.7426E+08	+	&8.9877E+08	&\bf1.2343E+05	&9.9662E+04\\
            \midrule
            F2	&2.3865E+04	+	&5.8286E+03	&9.6220E+03	+	&4.1287E+03	&8.2402E+03	+	&1.9775E+03	&5.6336E+04	+	&3.6873E+04	&9.6060E+03	+	&3.7013E+03	&3.9714E+03	=	&2.3480E+03	&\bf4.1265E+02	-	&1.9807E+02	&9.1584E+03	+	&3.8958E+03	&2.4170E+03	&8.1216E+02\\
            \midrule
            F3	&5.0311E+02	+	&2.5804E+01	&4.6819E+02	+	&5.0290E+01	&6.7225E+02	+	&1.1492E+02	&1.3279E+03	+	&4.8705E+02	&5.7053E+02	+	&8.0369E+01	&4.3907E+02	+	&3.7534E+01	&4.1230E+02	=	&1.9422E+01	&4.3971E+02	+	&5.2235E+01	&\bf4.1033E+02	&1.3291E+01\\
            \midrule
            F4	&5.5696E+02	+	&7.1677E+00	&5.6424E+02	+	&8.1432E+00	&5.7047E+02	+	&1.4301E+01	&5.9979E+02	+	&2.0055E+01	&5.6038E+02	+	&1.4649E+01	&5.4997E+02	+	&1.5834E+01	&5.3030E+02	=	&1.0172E+01	&5.4623E+02	+	&1.2385E+01	&\bf5.2629E+02	&5.4220E+00\\
            \midrule
            F5	&6.3068E+02	+	&4.8673E+00	&6.2510E+02	+	&9.4072E+00	&6.4093E+02	+	&1.0354E+01	&6.5849E+02	+	&1.5464E+01	&6.3281E+02	+	&9.4748E+00	&6.3596E+02	+	&1.0043E+01	&6.0850E+02	+	&6.4012E+00	&6.2502E+02	+	&9.9558E+00	&\bf6.0036E+02	&1.0642E-01\\
            \midrule
            F6	&8.6842E+02	+	&2.5513E+01	&7.8715E+02	+	&2.3756E+01	&8.3088E+02	+	&1.8065E+01	&8.6909E+02	+	&4.1911E+01	&7.9815E+02	+	&1.6787E+01	&8.0461E+02	+	&4.3847E+01	&7.4874E+02	=	&1.1291E+01	&7.6594E+02	+	&1.3538E+01	&\bf7.4423E+02	&5.6326E+00\\
            \midrule
            F7	&8.7036E+02	+	&8.2647E+00	&8.6639E+02	+	&1.0272E+01	&8.6095E+02	+	&8.5133E+00	&8.8601E+02	+	&1.4748E+01	&8.5262E+02	+	&1.1065E+01	&8.4070E+02	+	&1.3957E+01	&\bf8.2301E+02	-	&7.0388E+00	&8.3540E+02	+	&1.1641E+01	&8.2702E+02	&5.2658E+00\\
            \midrule
            F8	&2.1508E+03	+	&2.6092E+02	&1.2825E+03	+	&2.9553E+02	&1.6626E+03	+	&2.8972E+02	&2.2436E+03	+	&6.0884E+02	&1.3945E+03	+	&2.5411E+02	&1.6071E+03	+	&3.0445E+02	&1.0034E+03	+	&1.2054E+02	&1.1427E+03	+	&1.5922E+02	&\bf9.0071E+02	&4.9611E-01\\
            \midrule
            F9	&\bf1.8891E+03	-	&2.0177E+02	&2.7307E+03	+	&1.4971E+02	&2.4165E+03	=	&2.3897E+02	&3.1225E+03	+	&2.5338E+02	&2.7210E+03	+	&2.2631E+02	&2.1204E+03	-	&3.7867E+02	&1.9384E+03	-	&3.0102E+02	&2.3367E+03	=	&2.7781E+02	&2.3146E+03	&1.7654E+02\\
            \midrule
            F10	&1.1998E+03	+	&1.7809E+01	&1.2630E+03	+	&1.0348E+02	&3.2325E+03	+	&2.8591E+03	&6.3580E+03	+	&6.2999E+03	&1.4146E+03	+	&2.2254E+02	&1.2122E+03	+	&7.5409E+01	&1.1350E+03	+	&3.8138E+01	&1.3807E+03	+	&7.9993E+02	&\bf1.1142E+03	&4.8103E+00\\
            \midrule
            F11	&1.1584E+07	+	&4.6930E+06	&3.8881E+07	+	&5.3122E+07	&1.4298E+08	+	&1.5545E+08	&6.7902E+08	+	&4.4917E+08	&3.2316E+07	+	&3.7092E+07	&7.5518E+05	-	&1.2930E+06	&\bf3.0549E+05	-	&9.8597E+05	&4.4829E+06	+	&2.1912E+06	&8.5832E+05	&7.1459E+05\\
            \midrule
            F12	&\bf1.9706E+03	-	&2.0842E+02	&5.0561E+04	+	&8.3743E+04	&4.2770E+05	+	&6.5299E+05	&1.5592E+07	+	&2.3442E+07	&5.6004E+05	+	&6.8950E+05	&1.3138E+04	+	&9.5493E+03	&1.3016E+04	+	&1.0260E+04	&1.4237E+04	+	&7.8521E+03	&3.3032E+03	&1.2785E+03\\
            \midrule
            F13	&\bf1.4383E+03	-	&4.0188E+00	&2.3528E+03	+	&1.2499E+03	&1.8786E+03	+	&5.4471E+02	&5.9239E+03	+	&1.2837E+04	&3.2236E+03	+	&3.0739E+03	&1.5202E+03	+	&3.3492E+01	&1.6268E+03	+	&3.5839E+02	&3.9834E+03	+	&2.0029E+03	&1.4672E+03	&1.4005E+01\\
            \midrule
            F14	&\bf1.5935E+03	-	&1.8431E+01	&9.6860E+03	+	&1.0289E+04	&7.4483E+03	+	&4.7168E+03	&2.2371E+04	+	&1.3829E+04	&1.2171E+04	+	&7.9572E+03	&1.7259E+03	=	&1.2962E+02	&3.9340E+03	+	&3.2239E+03	&8.9633E+03	+	&7.0916E+03	&1.8502E+03	&2.1149E+02\\
            \midrule
            F15	&1.7286E+03	+	&5.6747E+01	&1.9264E+03	+	&1.1201E+02	&1.8983E+03	+	&1.0406E+02	&2.2413E+03	+	&1.9742E+02	&1.9452E+03	+	&1.1732E+02	&1.8797E+03	+	&1.7227E+02	&1.8006E+03	+	&1.4283E+02	&1.9058E+03	+	&1.3344E+02	&\bf1.6527E+03	&3.1736E+01\\
            \midrule
            F16	&1.7940E+03	+	&1.3243E+01	&1.8523E+03	+	&4.4471E+01	&1.8474E+03	+	&4.6642E+01	&2.0148E+03	+	&1.2623E+02	&1.8162E+03	+	&2.8135E+01	&1.8203E+03	+	&6.4346E+01	&\bf1.7513E+03	=	&2.0788E+01	&1.8169E+03	+	&5.2844E+01	&1.7561E+03	&1.1081E+01\\
            \midrule
            F17	&5.5629E+03	+	&1.8116E+03	&1.0757E+06	+	&1.8108E+06	&1.6556E+06	+	&2.1642E+06	&1.6170E+07	+	&2.4147E+07	&4.8387E+05	+	&1.0896E+06	&1.4072E+04	=	&1.3977E+04	&2.0071E+04	+	&1.6981E+04	&3.9175E+04	+	&2.3285E+04	&\bf4.3994E+03	&1.5618E+03\\
            \midrule
            F18	&\bf1.9422E+03	-	&1.0348E+01	&1.6701E+04	+	&1.3618E+04	&2.5565E+04	+	&2.5571E+04	&2.5765E+06	+	&3.8723E+06	&1.2808E+04	+	&1.6325E+04	&1.3629E+04	+	&2.0289E+04	&8.9783E+03	+	&9.2360E+03	&1.7628E+04	+	&5.4216E+04	&2.0131E+03	&1.3699E+02\\
            \midrule
            F19	&2.0834E+03	+	&1.4789E+01	&2.1587E+03	+	&2.8211E+01	&2.1632E+03	+	&5.7575E+01	&2.3499E+03	+	&7.3892E+01	&2.1705E+03	+	&5.2441E+01	&2.1954E+03	+	&7.2114E+01	&2.0528E+03	=	&4.1794E+01	&2.1778E+03	+	&7.8325E+01	&\bf2.0487E+03	&1.3623E+01\\
            \midrule
            F20	&2.3351E+03	+	&4.8258E+01	&\bf2.2419E+03	-	&2.7355E+01	&2.3084E+03	=	&5.3615E+01	&2.3886E+03	+	&2.7493E+01	&2.2770E+03	=	&5.2571E+01	&2.3218E+03	+	&4.2134E+01	&2.2902E+03	=	&5.7281E+01	&2.3105E+03	+	&5.0153E+01	&2.2812E+03	&4.3785E+01\\
            \midrule
            F21	&2.4376E+03	+	&4.7796E+01	&2.3417E+03	+	&4.7417E+01	&2.5960E+03	+	&1.8894E+02	&3.1021E+03	+	&4.4076E+02	&2.4861E+03	+	&1.2859E+02	&2.3531E+03	+	&4.1124E+01	&2.3052E+03	+	&1.7881E+01	&2.3383E+03	+	&4.7855E+01	&\bf2.3051E+03	&9.8765E+00\\
            \midrule
            F22	&2.6497E+03	+	&6.8711E+00	&2.6650E+03	+	&1.1657E+01	&2.6613E+03	+	&2.3911E+01	&2.7259E+03	+	&4.0017E+01	&2.6690E+03	+	&1.8773E+01	&2.6650E+03	+	&2.9100E+01	&2.6482E+03	+	&1.8087E+01	&2.7207E+03	+	&4.4588E+01	&\bf2.6251E+03	&4.5071E+00\\
            \midrule
            F23	&2.7839E+03	+	&7.2568E+00	&2.7578E+03	+	&5.8372E+01	&2.7644E+03	+	&6.2466E+01	&2.8763E+03	+	&4.8660E+01	&2.7523E+03	+	&8.0621E+01	&2.7773E+03	+	&7.9699E+01	&2.7422E+03	=	&9.5992E+01	&2.7857E+03	+	&7.9358E+01	&\bf2.7222E+03	&6.3011E+01\\
            \midrule
            F24	&2.9963E+03	+	&1.3771E+01	&2.9717E+03	+	&3.0679E+01	&3.1361E+03	+	&1.3151E+02	&3.6633E+03	+	&3.9878E+02	&3.0798E+03	+	&8.1302E+01	&2.9649E+03	+	&4.9162E+01	&\bf2.9347E+03	=	&2.9435E+01	&2.9663E+03	+	&5.5684E+01	&2.9394E+03	&1.6029E+01\\
            \midrule
            F25	&3.1024E+03	+	&2.9921E+01	&3.1513E+03	+	&1.2244E+02	&3.4113E+03	+	&2.0207E+02	&4.2273E+03	+	&4.8243E+02	&3.2919E+03	+	&1.1966E+02	&3.4701E+03	+	&3.8346E+02	&3.1103E+03	+	&2.4965E+02	&3.4031E+03	+	&3.2412E+02	&\bf2.9322E+03	&1.4842E+01\\
            \midrule
            F26	&3.0964E+03	=	&1.7936E+00	&3.1240E+03	+	&1.0471E+01	&3.1508E+03	+	&3.7136E+01	&3.2700E+03	+	&6.5192E+01	&3.1415E+03	+	&2.7283E+01	&3.1806E+03	+	&4.7025E+01	&3.1064E+03	+	&2.1802E+01	&3.1738E+03	+	&3.5527E+01	&\bf3.0960E+03	&2.4837E+00\\
            \midrule
            F27	&\bf3.2243E+03	=	&3.5447E+01	&3.3304E+03	=	&7.3201E+01	&3.6534E+03	+	&1.7846E+02	&3.8132E+03	+	&1.5304E+02	&3.3897E+03	+	&1.1161E+02	&3.3892E+03	=	&1.7945E+02	&3.2966E+03	=	&1.1887E+02	&3.4236E+03	+	&1.3991E+02	&3.2807E+03	&9.8800E+01\\
            \midrule
            F28	&3.2527E+03	+	&3.4048E+01	&3.3179E+03	+	&4.0784E+01	&3.2942E+03	+	&6.3533E+01	&3.5899E+03	+	&1.4690E+02	&3.3451E+03	+	&6.6558E+01	&3.3731E+03	+	&1.1997E+02	&3.2466E+03	=	&6.4696E+01	&3.2964E+03	+	&6.9154E+01	&\bf3.2137E+03	&3.1133E+01\\
            \midrule
            F29	&6.4807E+05	=	&3.2173E+05	&4.3263E+06	+	&2.2985E+06	&4.4027E+06	+	&4.1205E+06	&2.2141E+07	+	&1.5885E+07	&4.1837E+06	+	&3.8104E+06	&9.6681E+05	=	&1.0951E+06	&\bf5.7073E+05	=	&6.5496E+05	&8.5264E+05	=	&1.3208E+06	&7.2994E+05	&8.0763E+05\\
            \midrule
             $ $ &21/3/5			 &&27/1/1			&&27/2/0			&&29/0/0			&&28/1/0			&&22/5/2			&&14/11/4			&&27/2/0\\		
		\end{tabular}
}
\end{sidewaystable*}

\begin{sidewaystable*}[thp]
	\scriptsize
	\centering
	\renewcommand\arraystretch{1.2}
	\caption{The mean and standard deviation of the optimal solutions obtained from 30 trial runs for all algorithms are reported for the 30-D CEC2017 function functions. The meaning of all symbols is the same as Tab. \ref{tbl:CEC2017_10D}.}
	\label{tbl:CEC2017_30D}
	\resizebox{\linewidth}{!}{
		\begin{tabular}{cccccccccccccccccccccc}
		  \toprule
			\multirow{3}{*}{Func} & \multicolumn{2}{c}{DE} & \multicolumn{2}{c}{PSO} & \multicolumn{2}{c}{AO} & \multicolumn{2}{c}{SOA} &
            \multicolumn{2}{c}{SFO} & 
            \multicolumn{2}{c}{WOA} & 
            \multicolumn{2}{c}{HBA} & 
            \multicolumn{2}{c}{TSA} & 
            \multicolumn{2}{c}{MBGO}
   \\ \\
			\cmidrule(r){2-3} \cmidrule(r){4-5} \cmidrule(r){6-7} \cmidrule(r){8-9} \cmidrule(r){10-11} \cmidrule(r){12-13} \cmidrule(r){14-15} \cmidrule(r){16-17} \cmidrule(r){18-19} 
                & mean & std & mean & std & mean & std & mean & std
                & mean & std & mean & std
                & mean & std & mean & std
                & mean & std
                \\\bottomrule
            \midrule
            F1	&3.8728E+10	+	&5.0609E+09	&4.3502E+09	+	&1.8502E+09	&4.9914E+10	+	&5.6187E+09	&6.0064E+10	+	&8.1038E+09	&4.3065E+10	+	&5.7235E+09	&8.0108E+09	+	&3.3661E+09	&4.4222E+09	+	&2.9388E+09	&2.1478E+10	+	&5.2294E+09&\bf 3.9999E+05	&2.1338E+05\\
            \midrule
            F2	&2.5069E+05	+	&3.1990E+04	&1.0906E+05	+	&1.9972E+04	&6.9965E+04	+	&1.0313E+04	&2.0419E+06	+	&4.3544E+06	&1.0838E+05	+	&1.7436E+04	&1.5208E+05	+	&7.6587E+04&\bf 2.6418E+04	-	&7.1502E+03	&1.2091E+05	+	&2.8951E+04	&3.7025E+04	&7.1681E+03\\
            \midrule
            F3	&3.5658E+03	+	&6.3634E+02	&2.3855E+03	+	&1.4994E+03	&1.1904E+04	+	&2.9191E+03	&1.8700E+04	+	&2.5700E+03	&9.4554E+03	+	&2.5533E+03	&1.0816E+03	+	&4.6614E+02	&7.8547E+02	+	&3.8188E+02	&2.5350E+03	+	&1.0029E+03&\bf 5.1967E+02	&2.6220E+01\\
            \midrule
            F4	&8.9886E+02	+	&2.2777E+01	&8.4437E+02	+	&4.1589E+01	&9.3907E+02	+	&4.7082E+01	&9.7528E+02	+	&5.5841E+01	&8.7332E+02	+	&3.1116E+01	&8.0271E+02	+	&5.9541E+01	&7.0232E+02	+	&2.9969E+01	&8.0449E+02	+	&4.4030E+01&\bf 6.6638E+02	&1.2301E+01\\
            \midrule
            F5	&6.7420E+02	+	&5.1205E+00	&6.5655E+02	+	&1.3181E+01	&6.8761E+02	+	&8.9043E+00	&6.9961E+02	+	&1.0795E+01	&6.7470E+02	+	&9.9508E+00	&6.6664E+02	+	&1.2440E+01	&6.5002E+02	+	&7.8441E+00	&6.6965E+02	+	&1.1006E+01&\bf 6.0019E+02	&1.6504E-01\\
            \midrule
            F6	&2.4298E+03	+	&1.8956E+02	&1.1467E+03	+	&7.9555E+01	&1.4626E+03	+	&6.2084E+01	&1.5082E+03	+	&6.2994E+01	&1.3970E+03	+	&5.4681E+01	&1.2780E+03	+	&8.8076E+01	&1.1004E+03	+	&5.9757E+01	&1.3420E+03	+	&9.1824E+01&\bf 9.1531E+02	&1.3879E+01\\
            \midrule
            F7	&1.1898E+03	+	&1.5408E+01	&1.1010E+03	+	&4.1053E+01	&1.1669E+03	+	&3.0372E+01	&1.1868E+03	+	&3.7024E+01	&1.1200E+03	+	&2.4730E+01	&1.0448E+03	+	&6.9744E+01	&9.6241E+02	=	&2.6398E+01	&1.0793E+03	+	&3.6819E+01&\bf 9.6197E+02	&1.4108E+01\\
            \midrule
            F8	&1.7332E+04	+	&1.5091E+03	&9.7535E+03	+	&3.5271E+03	&1.2070E+04	+	&1.7008E+03	&1.6444E+04	+	&3.8922E+03	&9.9746E+03	+	&1.8243E+03	&8.6427E+03	+	&2.8458E+03	&6.3008E+03	+	&1.1068E+03	&1.0233E+04	+	&2.3811E+03&\bf 9.1518E+02	&1.3926E+01\\
            \midrule
            F9	&8.2699E+03	+	&3.2513E+02	&8.7280E+03	+	&3.1049E+02	&8.2350E+03	+	&4.7279E+02	&9.5814E+03	+	&3.6381E+02	&8.8911E+03	+	&4.0856E+02	&6.6912E+03	-	&1.0825E+03&\bf 5.9299E+03	-	&7.9984E+02	&7.9628E+03	=	&5.9328E+02	&7.8569E+03	&3.6766E+02\\
            \midrule
            F10	&3.1316E+03	+	&3.7149E+02	&3.9701E+03	+	&1.7100E+03	&9.4367E+03	+	&2.0529E+03	&3.4245E+04	+	&1.4526E+04	&1.0053E+04	+	&2.8364E+03	&3.5149E+03	+	&1.7516E+03	&1.3464E+03	=	&7.6694E+01	&5.0166E+03	+	&1.7138E+03&\bf 1.3131E+03	&4.1529E+01\\
            \midrule
            F11	&1.7211E+09	+	&3.0898E+08	&5.1290E+08	+	&6.8377E+08	&8.5477E+09	+	&3.1290E+09	&1.4300E+10	+	&3.4107E+09	&6.2984E+09	+	&2.5569E+09	&1.3437E+08	+	&2.6821E+08	&2.7132E+07	+	&2.5119E+07	&1.2574E+09	+	&8.8895E+08&\bf 1.7308E+06	&1.0286E+06\\
            \midrule
            F12	&1.1030E+08	+	&3.7965E+07	&4.2866E+07	+	&1.4898E+08	&2.2078E+09	+	&9.9423E+08	&8.2170E+09	+	&4.6757E+09	&1.4042E+09	+	&9.4096E+08	&5.9220E+05	+	&2.0288E+06	&1.2119E+05	+	&1.0974E+05	&8.7798E+07	+	&1.2505E+08&\bf 2.7476E+04	&7.0999E+04\\
            \midrule
            F13	&1.0532E+04	+	&4.7461E+03	&5.6162E+05	+	&5.4288E+05	&1.3687E+06	+	&1.9843E+06	&1.7254E+07	+	&1.3161E+07	&1.8654E+06	+	&1.4949E+06	&4.2490E+05	+	&4.8760E+05	&6.4604E+04	+	&4.6451E+04	&8.6265E+05	+	&7.8873E+05&\bf 5.2153E+03	&2.8781E+03\\
            \midrule
            F14	&4.1373E+05	+	&1.5754E+05	&5.5905E+06	+	&1.4129E+07	&4.1178E+08	+	&2.8486E+08	&1.9920E+09	+	&1.2089E+09	&1.9891E+08	+	&1.2191E+08	&4.2748E+04	+	&2.9708E+04&\bf 1.2520E+04	=	&1.1489E+04	&3.5921E+06	+	&4.5400E+06	&1.4880E+04	&1.0204E+04\\
            \midrule
            F15	&3.9142E+03	+	&2.1528E+02	&4.2880E+03	+	&2.9930E+02	&4.4745E+03	+	&4.9327E+02	&6.1788E+03	+	&1.1618E+03	&4.4340E+03	+	&4.7161E+02	&3.7990E+03	+	&8.4047E+02&\bf 2.9542E+03	=	&3.8719E+02	&3.7050E+03	+	&5.0727E+02	&2.9714E+03	&2.4376E+02\\
            \midrule
            F16	&2.6723E+03	+	&1.6153E+02	&2.9664E+03	+	&2.1927E+02	&3.3736E+03	+	&2.9292E+02	&5.9779E+03	+	&4.2243E+03	&2.9951E+03	+	&2.2103E+02	&2.7514E+03	+	&3.0910E+02	&2.3888E+03	+	&2.8263E+02	&2.4858E+03	+	&2.3440E+02&\bf 2.0509E+03	&9.2353E+01\\
            \midrule
            F17	&4.7866E+06	+	&2.8191E+06	&1.1883E+07	+	&1.0894E+07	&2.5579E+07	+	&2.2746E+07	&2.2080E+08	+	&2.3998E+08	&2.0746E+07	+	&1.7701E+07	&1.5508E+06	+	&2.1712E+06	&8.6553E+05	+	&8.8257E+05	&5.5399E+06	+	&6.6024E+06&\bf 1.3189E+05	&6.1611E+04\\
            \midrule
            F18	&8.3228E+06	+	&3.1182E+06	&2.0227E+07	+	&5.6070E+07	&5.3923E+08	+	&2.8832E+08	&1.8328E+09	+	&9.2788E+08	&2.3078E+08	+	&1.7731E+08	&1.0095E+05	+	&1.4567E+05&\bf 1.3702E+04	=	&1.3995E+04	&6.0173E+06	+	&5.0015E+06	&1.7962E+04	&1.6741E+04\\
            \midrule
            F19	&2.6174E+03	=	&1.7028E+02	&2.9516E+03	+	&9.5156E+01	&2.8554E+03	+	&1.5177E+02	&3.3609E+03	+	&2.2337E+02	&3.0129E+03	+	&1.5627E+02	&2.9047E+03	+	&1.7810E+02	&2.5596E+03	=	&1.7195E+02	&2.7934E+03	+	&1.8604E+02&\bf 2.5338E+03	&9.8394E+01\\
            \midrule
            F20	&2.6649E+03	+	&1.7781E+01	&2.6027E+03	+	&3.8054E+01	&2.6899E+03	+	&4.4773E+01	&2.7808E+03	+	&5.7305E+01	&2.6450E+03	+	&3.5083E+01	&2.5796E+03	+	&5.5725E+01	&2.5021E+03	+	&3.6330E+01	&2.6012E+03	+	&3.3133E+01&\bf 2.4611E+03	&1.5175E+01\\
            \midrule
            F21	&1.0028E+04	+	&2.8090E+02	&4.1288E+03	+	&1.7673E+03	&9.3437E+03	+	&1.0607E+03	&1.0597E+04	+	&9.3450E+02	&7.2491E+03	+	&1.3597E+03	&7.7189E+03	+	&1.1137E+03	&5.9714E+03	+	&1.9898E+03	&8.0186E+03	+	&2.2926E+03&\bf 2.3068E+03	&2.9102E+00\\
            \midrule
            F22	&2.9854E+03	+	&2.1517E+01	&3.0835E+03	+	&7.1539E+01	&3.2495E+03	+	&1.1743E+02	&3.5254E+03	+	&1.1898E+02	&3.2886E+03	+	&9.9796E+01	&3.2075E+03	+	&1.0923E+02	&3.0496E+03	+	&7.4218E+01	&3.3827E+03	+	&1.3219E+02&\bf 2.8168E+03	&2.3026E+01\\
            \midrule
            F23	&3.1281E+03	+	&1.6360E+01	&3.2377E+03	+	&6.2597E+01	&3.3608E+03	+	&1.0822E+02	&3.7525E+03	+	&3.0598E+02	&3.4290E+03	+	&9.6420E+01	&3.3091E+03	+	&1.4969E+02	&3.2291E+03	+	&1.0210E+02	&3.5035E+03	+	&1.0953E+02&\bf 2.9955E+03	&1.5911E+01\\
            \midrule
            F24	&6.4496E+03	+	&7.0250E+02	&3.2180E+03	+	&1.6632E+02	&4.5885E+03	+	&3.9764E+02	&5.8198E+03	+	&5.9821E+02	&4.6837E+03	+	&4.4923E+02	&3.1393E+03	+	&8.3889E+01	&3.0143E+03	+	&6.5582E+01	&3.4229E+03	+	&2.5761E+02&\bf 2.9206E+03	&1.9850E+01\\
            \midrule
            F25	&7.5970E+03	+	&2.6663E+02	&8.2050E+03	+	&5.5717E+02	&1.0463E+04	+	&1.1581E+03	&1.2011E+04	+	&1.3815E+03	&9.8868E+03	+	&9.9612E+02	&8.3476E+03	+	&1.0505E+03	&7.0872E+03	+	&1.0680E+03	&9.1372E+03	+	&1.3384E+03&\bf 5.0594E+03	&6.9290E+02\\
            \midrule
            F26	&3.2813E+03	+	&1.7358E+01	&3.6160E+03	+	&1.1583E+02	&3.7535E+03	+	&2.1358E+02	&4.4360E+03	+	&3.7989E+02	&3.8136E+03	+	&1.9701E+02	&3.6145E+03	+	&2.3862E+02	&3.3544E+03	+	&8.6376E+01	&3.6017E+03	+	&1.2673E+02&\bf 3.2291E+03	&1.1674E+01\\
            \midrule
            F27	&4.7900E+03	+	&7.3332E+02	&4.1294E+03	+	&7.5415E+02	&7.2253E+03	+	&6.5758E+02	&7.6238E+03	+	&8.1317E+02	&6.2442E+03	+	&6.0910E+02	&3.8180E+03	+	&2.8034E+02	&3.3960E+03	+	&7.5676E+01	&4.4605E+03	+	&3.9363E+02&\bf 3.2962E+03	&3.3187E+01\\
            \midrule
            F28	&4.3526E+03	+	&3.1228E+02	&5.3329E+03	+	&3.5824E+02	&5.8493E+03	+	&7.6345E+02	&9.0717E+03	+	&2.3637E+03	&5.7024E+03	+	&5.3510E+02	&5.4865E+03	+	&5.0940E+02	&4.4447E+03	+	&3.3059E+02	&5.1180E+03	+	&3.8691E+02&\bf 3.9008E+03	&1.6400E+02\\
            \midrule
            F29	&1.0604E+07	+	&3.9765E+06	&1.9096E+07	+	&2.9057E+07	&5.5429E+08	+	&2.3612E+08	&1.7611E+09	+	&1.1731E+09	&3.1372E+08	+	&1.9024E+08	&2.0572E+06	+	&3.2932E+06	&5.7292E+05	+	&5.0891E+05	&3.6931E+07	+	&3.6575E+07&\bf 5.3820E+04	&2.9038E+04\\
            \midrule
            $ $ &28/1/0		&	&29/0/0		&	&29/0/0		&	&29/0/0		&	&29/0/0		&	&28/0/1		&	 &21/6/2	&		&28/1/0\\		
		\end{tabular}
}
\end{sidewaystable*}

\begin{sidewaystable*}[thp]
	\scriptsize
	\centering
	\renewcommand\arraystretch{1.2}
	\caption{The mean and standard deviation of the optimal solutions obtained from 30 trial runs for all algorithms are reported for the 50-D CEC2020 function functions. The meaning of all symbols is the same as Tab. \ref{tbl:CEC2017_10D}.}
	\label{tbl:CEC2020_50D}
	\resizebox{\linewidth}{!}{
		\begin{tabular}{cccccccccccccccccccccc}
		  \toprule
			\multirow{3}{*}{Func} & \multicolumn{2}{c}{DE} & \multicolumn{2}{c}{PSO} & \multicolumn{2}{c}{AO} & \multicolumn{2}{c}{SOA} &
            \multicolumn{2}{c}{SFO} & 
            \multicolumn{2}{c}{WOA} & 
            \multicolumn{2}{c}{HBA} & 
            \multicolumn{2}{c}{TSA} & 
            \multicolumn{2}{c}{MBGO}
   \\ \\
			\cmidrule(r){2-3} \cmidrule(r){4-5} \cmidrule(r){6-7} \cmidrule(r){8-9} \cmidrule(r){10-11} \cmidrule(r){12-13} \cmidrule(r){14-15} \cmidrule(r){16-17} \cmidrule(r){18-19}  
                & mean & std & mean & std & mean & std & mean & std
                & mean & std & mean & std
                & mean & std & mean & std
                & mean & std
                \\\bottomrule
            \midrule
            F1	&1.0735E+11	&1.0424E+10	+	&1.9074E+10	&3.7568E+09	+	&9.6145E+10	&7.3230E+09	+	&1.1696E+11	&7.3465E+09	+	&9.8024E+10	&8.5482E+09	+	&2.1821E+10	&6.8594E+09	+	&1.7790E+10	&7.1328E+09	+	&7.2565E+10	&8.9100E+09	+	&\bf1.1270E+04	&7.0292E+03\\
            \midrule
            F2	&1.1506E+13	&1.3854E+12	+	&2.3224E+12	&5.4535E+11	+	&1.1529E+13	&8.3263E+11	+	&1.3659E+13	&1.0848E+12	+	&1.1463E+13	&1.2099E+12	+	&2.3227E+12	&7.6796E+11	+	&1.5718E+12	&6.0954E+11	+	&7.4400E+12	&1.0085E+12	+	&\bf3.2366E+06	&2.3897E+06\\
            \midrule
            F3	&4.1251E+12	&4.6035E+11	+	&6.7926E+11	&1.9848E+11	+	&4.0254E+12	&2.8703E+11	+	&4.9051E+12	&5.2138E+11	+	&3.8443E+12	&3.3507E+11	+	&7.4047E+11	&2.5709E+11	+	&4.9464E+11	&2.1317E+11	+	&2.3369E+12	&2.5821E+11	+	&\bf5.3442E+05	&3.9546E+05\\
            \midrule
            F4	&1.4032E+06	&6.3204E+05	+	&8.9186E+04	&1.3069E+05	+	&2.4741E+06	&7.5748E+05	+	&5.6961E+06	&2.2665E+06	+	&2.2050E+06	&9.6700E+05	+	&3.7353E+04	&3.3682E+04	+	&6.3055E+03	&6.9536E+03	+	&5.8974E+05	&4.2554E+05	+	&\bf1.9390E+03	&4.7258E+00\\
            \midrule
            F5	&6.2294E+07	&2.1771E+07	+	&6.6321E+07	&4.0643E+07	+	&3.4876E+08	&1.5992E+08	+	&8.2669E+08	&3.9760E+08	+	&2.3913E+08	&1.4098E+08	+	&1.0332E+07	&6.9870E+06	+	&2.9838E+06	&1.3050E+06	+	&2.9063E+07	&2.1271E+07	+	&\bf7.2323E+05	&3.5151E+05\\
            \midrule
            F6	&9.2213E+07	&3.2226E+07	+	&1.0339E+07	&1.4644E+07	+	&2.4749E+09	&1.7390E+09	+	&6.5706E+09	&3.0102E+09	+	&1.5094E+09	&1.0228E+09	+	&8.4905E+05	&1.0048E+06	+	&1.2390E+05	&2.3009E+05	+	&1.1285E+08	&2.1738E+08	+	&\bf6.0837E+03	&4.3520E+03\\
            \midrule
            F7	&7.5293E+08	&1.8051E+08	+	&9.1527E+08	&7.6454E+08	+	&7.9962E+09	&6.4126E+09	+	&2.4756E+10	&1.2943E+10	+	&5.8246E+09	&3.2105E+09	+	&3.5848E+07	&3.2109E+07	+	&9.1699E+06	&5.6497E+06	=	&5.9403E+08	&3.5267E+08	+	&\bf3.2943E+05	&1.9094E+05\\
            \midrule
            F8	&3.1886E+03	&1.0644E+02	+	&4.1366E+03	&9.3510E+02	+	&8.5875E+03	&2.5559E+03	+	&1.4182E+04	&1.7978E+03	+	&9.1301E+03	&1.5912E+03	+	&9.6843E+03	&2.3091E+03	+	&3.9288E+03	&8.9360E+02	+	&8.1978E+03	&2.4261E+03	+	&\bf2.5467E+03	&4.2374E+01\\
            \midrule
            F9	&2.7436E+04	&1.9050E+03	+	&2.3072E+04	&3.3678E+03	+	&6.1229E+04	&2.5016E+03	+	&6.9090E+04	&1.8918E+03	+	&6.2767E+04	&4.1168E+03	+	&3.3494E+04	&1.2397E+04	+	&2.2905E+04	&7.1212E+03	+	&4.8971E+04	&6.1176E+03	+	&\bf2.6600E+03	&1.5308E+02\\
            \midrule
            F10	&1.2274E+04	&1.7814E+03	+	&8.4358E+03	&1.8796E+03	+	&2.6724E+04	&3.5672E+03	+	&3.1082E+04	&5.6758E+03	+	&2.1609E+04	&3.3043E+03	+	&5.3565E+03	&7.4736E+02	+	&4.7492E+03	&5.6049E+02	+	&9.8201E+03	&1.4097E+03	+	&\bf3.0257E+03	&3.7477E+01\\
            \midrule
            $ $ &10/0/0			&&10/0/0			&&10/0/0			&&10/0/0			&&10/0/0			 &&10/0/0			&&10/0/0	&&10/0/0\\	
                		\end{tabular}
}
\end{sidewaystable*}

\begin{sidewaystable*}[thp]
	\scriptsize
	\centering
	\renewcommand\arraystretch{1.2}
	\caption{The mean and standard deviation of the optimal solutions obtained from 30 trial runs for all algorithms are reported for the 100-D CEC2020 function functions. The meaning of all symbols is the same as Tab. \ref{tbl:CEC2017_10D}.}
	\label{tbl:CEC2020_100D}
	\resizebox{\linewidth}{!}{
		\begin{tabular}{cccccccccccccccccccccc}
		  \toprule
			\multirow{3}{*}{Func} & \multicolumn{2}{c}{DE} & \multicolumn{2}{c}{PSO} & \multicolumn{2}{c}{AO} & \multicolumn{2}{c}{SOA} &
            \multicolumn{2}{c}{SFO} & 
            \multicolumn{2}{c}{WOA} & 
            \multicolumn{2}{c}{HBA} & 
            \multicolumn{2}{c}{TSA} & 
            \multicolumn{2}{c}{MBGO}
   \\ \\
			\cmidrule(r){2-3} \cmidrule(r){4-5} \cmidrule(r){6-7} \cmidrule(r){8-9} \cmidrule(r){10-11} \cmidrule(r){12-13} \cmidrule(r){14-15} \cmidrule(r){16-17} \cmidrule(r){18-19}  
                & mean & std & mean & std & mean & std & mean & std
                & mean & std & mean & std
                & mean & std & mean & std
                & mean & std
                \\\bottomrule
            \midrule
            F1	&3.3290E+11	&2.4114E+10	+	&1.0081E+11	&1.3299E+10	+	&2.4413E+11	&1.0715E+10	+	&2.7227E+11	&1.0621E+10	+	&2.5156E+11	&1.4181E+10	+	&7.6968E+10	&1.5254E+10	+	&6.7782E+10	&1.4357E+10	+	&2.2502E+11	&1.6060E+10	+	&\bf2.7756E+05	&2.6120E+05\\
            \midrule
            F2	&3.0822E+13	&2.1708E+12	+	&1.0952E+13	&1.3767E+12	+	&2.8573E+13	&1.2942E+12	+	&3.2128E+13	&1.3986E+12	+	&2.8883E+13	&1.4358E+12	+	&7.5722E+12	&1.7557E+12	+	&7.5186E+12	&1.6116E+12	+	&2.3821E+13	&2.0692E+12	+	&\bf3.6781E+07	&4.1426E+07\\
            \midrule
            F3	&1.1659E+13	&8.5723E+11	+	&3.6026E+12	&4.6601E+11	+	&1.0059E+13	&4.7191E+11	+	&1.1046E+13	&4.2657E+11	+	&9.9554E+12	&5.4303E+11	+	&2.8013E+12	&4.9218E+11	+	&2.6187E+12	&4.5279E+11	+	&8.1582E+12	&6.9120E+11	+	&\bf1.1970E+07	&1.3542E+07\\
            \midrule
            F4	&2.0954E+07	&6.6383E+06	+	&2.5096E+05	&1.0988E+05	+	&5.1398E+06	&1.3197E+06	+	&9.7188E+06	&2.5867E+06	+	&6.5366E+06	&1.8470E+06	+	&9.6745E+04	&5.3899E+04	+	&3.2934E+04	&1.8880E+04	+	&2.8721E+06	&1.0506E+06	+	&\bf1.9973E+03	&1.1235E+01\\
            \midrule
            F5	&7.6995E+08	&1.0394E+08	+	&5.8197E+08	&3.3919E+08	+	&1.7890E+09	&5.2409E+08	+	&2.8889E+09	&8.9001E+08	+	&1.0579E+09	&3.2079E+08	+	&5.2976E+07	&1.9440E+07	+	&3.3856E+07	&1.3806E+07	+	&3.3186E+08	&1.2524E+08	+	&\bf5.2490E+06	&1.6354E+06\\
            \midrule
            F6	&3.4297E+09	&1.2393E+09	+	&5.9416E+08	&7.6776E+08	+	&4.1778E+10	&1.0344E+10	+	&6.1234E+10	&1.6841E+10	+	&3.1922E+10	&8.8112E+09	+	&4.0633E+07	&8.4946E+07	+	&3.7576E+07	&8.6541E+07	+	&7.2745E+09	&3.1452E+09	+	&\bf7.9953E+03	&4.4773E+03\\
            \midrule
            F7	&5.2077E+09	&1.4777E+09	+	&4.4154E+09	&2.8476E+09	+	&3.4314E+10	&9.6404E+09	+	&4.9552E+10	&8.7095E+09	+	&2.4942E+10	&7.4644E+09	+	&1.4799E+08	&7.7045E+07	+	&5.6523E+07	&3.3044E+07	+	&4.8550E+09	&2.1885E+09	+	&\bf2.1715E+06	&1.0121E+06\\
            \midrule
            F8	&4.5090E+03	&3.7749E+02	+	&8.7368E+03	&1.7341E+03	+	&2.6462E+04	&3.9228E+03	+	&2.9316E+04	&1.4443E+03	+	&2.2976E+04	&2.6009E+03	+	&2.2923E+04	&3.6860E+03	+	&1.0981E+04	&4.1486E+03	+	&2.2774E+04	&4.5975E+03	+	&\bf2.9977E+03	&1.3547E+02\\
            \midrule
            F9	&1.1576E+05	&1.2683E+04	+	&1.0217E+05	&1.0158E+04	+	&1.7486E+05	&3.3467E+03	+	&1.8716E+05	&3.6725E+03	+	&1.7845E+05	&6.6147E+03	+	&1.1740E+05	&1.8212E+04	+	&9.1935E+04	&2.3643E+04	+	&1.7209E+05	&9.8243E+03	+	&\bf2.8337E+03	&3.5557E+02\\
            \midrule
            F10	&4.1929E+04	&7.8693E+03	+	&1.2357E+04	&1.5964E+03	+	&3.7320E+04	&3.4547E+03	+	&4.7240E+04	&5.1619E+03	+	&4.0707E+04	&4.7477E+03	+	&7.7985E+03	&9.9148E+02	+	&6.7834E+03	&8.5499E+02	+	&2.8233E+04	&4.1414E+03	+	&\bf3.3497E+03	&5.7358E+01\\
            \midrule
             $ $ &10/0/0			&&10/0/0			&&10/0/0			&&10/0/0			&&10/0/0			 &&10/0/0			&&10/0/0	&&10/0/0\\	
                		\end{tabular}
}
\end{sidewaystable*}

\begin{sidewaystable*}[thp]
\scriptsize
	\centering
	\renewcommand\arraystretch{1.2}
	\caption{The mean and standard deviation of the optimal solutions obtained from 30 trial runs for all algorithms are reported for real-world problems. The meaning of all symbols is the same as Tab. \ref{tbl:CEC2017_10D}.}
	\label{tbl:real-world problems results}
	\resizebox{\linewidth}{!}{
		\begin{tabular}{cccccccccccccccccccccc}
		  \toprule
			\multirow{3}{*}{Pro.} & \multicolumn{2}{c}{DE} & \multicolumn{2}{c}{PSO} & \multicolumn{2}{c}{AO} & \multicolumn{2}{c}{SOA} &
            \multicolumn{2}{c}{SFO} & 
            \multicolumn{2}{c}{WOA} & 
            \multicolumn{2}{c}{HBA} & 
            \multicolumn{2}{c}{TSA} & 
            \multicolumn{2}{c}{MBGO}
   \\ \\
			\cmidrule(r){2-3} \cmidrule(r){4-5} \cmidrule(r){6-7} \cmidrule(r){8-9} \cmidrule(r){10-11} \cmidrule(r){12-13} \cmidrule(r){14-15} \cmidrule(r){16-17} \cmidrule(r){18-19}  
                & mean & std & mean & std & mean & std & mean & std
                & mean & std & mean & std
                & mean & std & mean & std
                & mean & std
                \\\bottomrule
            \midrule
            WBP	&1.6835E+00	&3.0067E-04	+	&1.9511E+00	&8.3384E-02	+	&3.3361E+00	&8.1034E-01	+	&3.6966E+00	&1.0326E+00	+	&2.5693E+00	&2.6176E-01	+	&3.0570E+00	&7.3228E-01	+	&1.6880E+00	&1.0087E-02	+	&1.7064E+00	&8.6894E-03	+	&\bf1.6826E+00	&5.2327E-05\\
            \midrule
            PVP	&6.1934E+03	&2.5662E+03	+	&1.4818E+04	&1.9961E+03	+	&1.5474E+04	&3.8811E+04	+	&1.1031E+05	&1.3282E+05	+	&1.3813E+04	&3.5615E+03	+	&4.6781E+04	&6.6477E+04	+	&7.0640E+03	&2.3014E+03	+	&7.3435E+03	&2.0214E+03	+	&\bf3.0738E+03	&9.7672E+02\\
            \midrule
            TBTD	&\bf2.6390E+02	&0.0000E+00	-	&2.6410E+02	&1.6711E-01	+	&2.6458E+02	&6.9978E-01	+	&2.6727E+02	&3.2705E+00	+	&2.6443E+02	&4.8632E-01	+	&2.6693E+02	&3.3105E+00	+	&2.6390E+02	&2.8062E-04	+	&2.6391E+02	&1.1650E-02	+	&2.6390E+02	&4.3062E-06\\
            \midrule
            GTD	&5.8830E-12	&9.1747E-12	+	&2.0535E-09	&4.1111E-09	+	&2.1612E-05	&1.1094E-04	+	&7.4905E-08	&3.9155E-07	+	&7.2398E-08	&2.6200E-07	+	&\bf0.0000E+00	&0.0000E+00	-	&2.0925E-14	&8.2780E-14	+	&4.1313E-12	&6.1100E-12	+	&2.9586E-15	&6.5181E-15\\
            \midrule
            CBD	&1.3429E+00	&1.5533E-03	+	&1.9847E+00	&3.3174E-01	+	&1.4668E+00	&6.1088E-02	+	&3.2055E+00	&1.3280E+00	+	&1.4527E+00	&4.0714E-02	+	&3.9796E+00	&1.4077E+00	+	&1.3402E+00	&1.6425E-04	+	&1.3417E+00	&8.8442E-04	+	&\bf1.3400E+00	&2.8699E-05\\
            \midrule
            IBD	&1.7458E-04	&2.7535E-09	=	&1.8300E-04	&3.2423E-06	+	&2.4825E-04	&5.6822E-05	+	&2.0785E-04	&5.9981E-05	+	&1.7612E-04	&4.6968E-06	+	&1.9189E-04	&3.7302E-05	+	&1.7458E-04	&8.2252E-19	-	&1.7458E-04	&2.3959E-10	+	&\bf1.7458E-04	&5.4210E-20\\
            \midrule
            TCD	&3.0150E+01	&6.2596E-13	+	&3.0260E+01	&4.4741E-02	+	&3.1591E+01	&1.0706E+00	+	&3.2006E+01	&1.1871E+00	+	&3.0620E+01	&2.8987E-01	+	&3.1237E+01	&1.0565E+00	+	&3.0150E+01	&3.8918E-15	-	&3.0174E+01	&1.1274E-02	+	&\bf3.0150E+01	&3.8896E-14\\
            \midrule
            PLD	&1.0574E+00	&3.8357E-11	-	&6.6202E+01	&4.3778E+01	+	&2.5952E+02	&2.3194E+02	+	&6.1817E+02	&5.2938E+02	+	&1.2909E+02	&1.2509E+02	+	&2.9641E+02	&2.1493E+02	+	&5.0982E+01	&7.6261E+01	+	&6.8014E+01	&8.1964E+01	+	&\bf1.0574E+00	&7.3849E-09\\
            \midrule
            CBHD	&6.8433E+00	&1.3660E-04	+	&7.2990E+00	&2.1369E-01	+	&8.9934E+00	&1.0649E+00	+	&9.2946E+00	&1.4089E+00	+	&7.3697E+00	&1.5462E-01	+	&7.2195E+00	&2.7971E-01	+	&6.8484E+00	&1.5198E-02	+	&6.8963E+00	&1.3338E-02	+	&\bf6.8430E+00	&1.9451E-11\\
            \midrule
            RCB	&1.6662E+02	&6.0548E-01	+	&1.5973E+02	&3.3297E-01	-	&1.7110E+02	&3.5498E+00	+	&1.7256E+02	&6.0799E+00	+	&1.6483E+02	&2.7695E+00	+	&1.6826E+02	&2.6540E+00	+	&\bf1.5986E+02	&1.8499E+00	-	&1.5956E+02	&1.9754E-01	-	&1.6061E+02	&1.1718E+00\\
            \midrule
             &7/1/2			 &&9/0/1			 &&10/0/0			 &&10/0/0			 &&10/0/0			  &&9/0/1			 &&7/0/3			  &&9/0/1		
                		\end{tabular}
}
\end{sidewaystable*}

\end{document}